\documentclass{article}

\usepackage[preprint]{neurips_2025}
\workshoptitle{Privacy-Preserving Machine Learning}

\usepackage[utf8]{inputenc} 
\usepackage[T1]{fontenc}    
\usepackage{hyperref}       
\usepackage{url}            
\usepackage{booktabs}       
\usepackage{amsfonts}       
\usepackage{amsmath}
\usepackage{nicefrac}       
\usepackage{microtype}      
\usepackage{xcolor}         
\usepackage{graphicx}
\usepackage{subcaption}
\usepackage{multirow}
\usepackage{wrapfig}
\usepackage{amsthm}
\usepackage{bm}

\newtheorem{prop}{Proposition}

\title{Privacy Preserving Diffusion Models for Mixed-Type Tabular Data Generation}

\author{
  Timur Sattarov\thanks{Research conducted in part as a PhD candidate at the School of Computer Science, University of St.Gallen (HSG), St.Gallen, Switzerland.} \\
  Deutsche Bundesbank\\
  Frankfurt am Main, Germany \\
  \texttt{timur.sattarov@bundesbank.de} \\
  \And
  Marco Schreyer \\
  Swiss Federal Audit Office \\
  Bern, Switzerland \\
  \texttt{marco.schreyer@efk.admin.ch} \\
  \AND
  Damian Borth \\
  University of St.Gallen \\
  St.Gallen, Switzerland \\
  \texttt{damian.borth@unisg.ch} \\
}

\begin{document}

\maketitle

\begin{abstract}
We introduce \textit{DP-FinDiff}, a differentially private diffusion framework for synthesizing mixed-type tabular data. \textit{DP-FinDiff} employs embedding-based representations for categorical features, reducing encoding overhead and scaling to high-dimensional datasets. To adapt DP-training to the diffusion process, we propose two privacy-aware training strategies: an \textit{adaptive timestep sampler} that aligns updates with diffusion dynamics, and a \textit{feature-aggregated loss} that mitigates clipping-induced bias. Together, these enhancements improve fidelity and downstream utility without weakening privacy guarantees. On financial and medical datasets, \textit{DP-FinDiff} achieves 16–42\% higher utility than DP baselines at comparable privacy levels, demonstrating its promise for safe and effective data sharing in sensitive domains.
\end{abstract}

\section{Introduction}

Tabular data lies at the core of high-stakes applications in domains such as finance and healthcare, where it enables policy design, risk modeling, and medical decision support. While synthetic data generation has emerged as a promising tool for data sharing and innovation, the sensitive nature of individual records raises serious privacy concerns. Without formal guarantees, generative models may still leak information through membership inference~\cite{shokri2017membership, salem2018} or model extraction attacks~\cite{carlini2021extracting, carlini2023extracting}. 

Differential Privacy (DP)~\citep{dwork2006our} provides a framework to limit such leakage by bounding the influence of each record during training. However, applying DP to tabular generative models poses key challenges: (i) \textit{encoding overhead} leads to large memory overhead and sparsity due to the high cardinality of one-hot encoded variables, (ii) \textit{scalability} slows training substantially as dataset size increases, (iii) \textit{utility loss} degrades when DP noise overwhelms fine-grained tabular patterns. These limitations hinder the deployment of private generative models in real-world, high-stakes settings. 

Diffusion probabilistic models (DDPMs)~\citep{sohl2015deep, ho2020denoising} have recently advanced at modeling mixed-type tabular data~\cite{kotelnikov2023tabddpm, sattarov2023findiff}, making them promising for sensitive data synthesis. However, combining them with DP is challenging: DP mechanisms inject noise into training, while diffusion models rely on precise denoising dynamics, which degrades sample fidelity.

\begin{figure}[t!]
  \centering
  \includegraphics[width=0.99\linewidth, ]{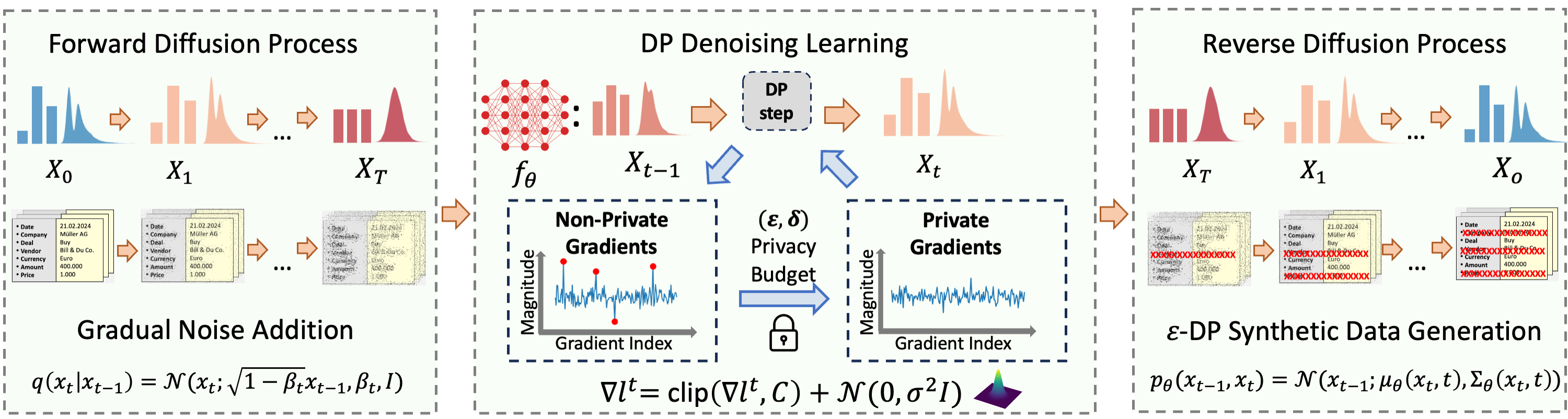}
    \caption{Schematic of the diffusion process for mixed-type tabular data with DP. 
    Left: forward noise addition from $X_0$ to $X_T$. 
    Middle: DP denoising learning with clipped, noised gradients under budget $(\varepsilon,\delta)$. 
    Right: reverse reconstruction from $X_T$ to $X_0$, producing $(\varepsilon,\delta)$-DP compliant synthetic data.}
  \label{fig:DP-FinDiff_model}
  \vspace{-0.3cm}
\end{figure}

To address these challenges, we propose \textit{DP-FinDiff},\footnote{The source code is available at: \url{https://github.com/sattarov/DP-FinDiff}} a conditional diffusion framework for mixed-type tabular data with formal $(\varepsilon, \delta)$-DP guarantees. \textit{DP-FinDiff} employs embeddings for categorical variables instead of one-hot encodings, reducing \textit{encoding overhead} and improving \textit{scalability}. To mitigate \textit{utility loss}, our method introduces two key advances: (i) a \textit{feature-aggregated loss} that reduces gradient variance and stabilizes DP-SGD, and (ii) an \textit{adaptive timestep sampler} that reallocates training to diffusion steps more robust to DP noise. Our main contributions are:

\begin{enumerate}
    \item \textbf{Methodology:} A conditional diffusion-based generative model \textit{DP-FinDiff} designed for mixed-type tabular data with $(\varepsilon,\delta)$-differential privacy guarantees.
    \item \textbf{Advancements:} Two privacy-aware enhancements \textit{feature-aggregated loss} and \textit{adaptive timestep sampling} that improve model utility and fidelity while preserving privacy.
    \item \textbf{Evaluation:} DP versions of TVAE, CTGAN, and TabDDPM as well as extensive experiments on financial and healthcare datasets, showing significant gains of \textit{DP-FinDiff}.
\end{enumerate}

\section{Related Work}

\textbf{Generative models for tabular data.}  
Recent surveys~\cite{fonseca2023tabular, kim2024generative} provide comprehensive overviews of generative modeling techniques for tabular data. Early work adapted GANs and VAEs to mixed-type tabular data, e.g., \textit{TVAE} and \textit{CTGAN}~\cite{tvae_ctgan}, CTAB-GAN~\cite{zhao2021ctab}, and Wasserstein GAN~\cite{engelmann2021conditional}. DoppelGANger~\cite{lin2019generating} targeted time-series synthesis, with follow-up work highlighting susceptibility to membership inference attacks~\cite{lin2021privacy}. More recently, growing line of research has shifted towards diffusion models~\cite{cao2022survey, li2025diffusion}, which have demonstrated strong performance in tabular synthesis~\cite{kotelnikov2023tabddpm, lee2023codi, zhang2023mixed, suh2023autodiff}. In particular, \textit{FinDiff}~\cite{sattarov2023findiff} demonstrated that categorical embeddings reduce sparsity and scale to high-cardinality domains. Extensions have targeted entity-awareness~\cite{liu2024entity} and imbalance~\cite{schreyer2024imb}, underscoring the flexibility of diffusion-based tabular generators.  

\vspace{-0.1cm}

\textbf{Differentially private generative models.}
A parallel stream of research studies DP-enabled generative modeling~\cite{fan2020survey, hassan2023deep, liu2024generative}, where DP mechanisms have been applied to GANs and VAEs~\cite{jordon2018pate, fang2022dp, kunar2021dtgan}, with applications in finance and healthcare~\cite{byrd2020differentially, zheng2020pcal, schreyer2022federated}. Recent work explored DP in score-based diffusion~\cite{truda2023generating, guan2024generating}, but these methods operate on continuous domains and do not address mixed-type tabular data. Moreover, most DP adaptations still rely on one-hot encodings, which introduce encoding overhead and degrade scalability under DP-SGD.  

\vspace{-0.1cm}

\textbf{Research gap.}  
To our knowledge, no prior work adapts the \emph{training dynamics} of diffusion models to differential privacy. Existing DP models either add noise without accounting for the diffusion process or suffer from high-dimensional encodings. Our work addresses this gap by introducing \emph{DP-aware training strategies} (adaptive timestep sampling, feature-aggregated loss) within a diffusion framework for mixed-type tabular data, achieving both formal guarantees and improved utility.

\section{Methodology}

\textit{DP-FinDiff} combines: (i) diffusion-based tabular synthesis with embeddings via \textit{FinDiff}~\cite{sattarov2023findiff},
(ii) DP-SGD with gradient clipping and Gaussian noise, and  
(iii) two training enhancements \emph{Adaptive Timestep (AT) sampling} and \emph{Feature-Aggregated (FA) loss} to mitigate noise-driven degradation.

\subsection{Background} 

\noindent\textbf{Diffusion models.}  
We adopt \textit{Denoising Diffusion Probabilistic Models} (DDPMs)~\cite{sohl2015deep,ho2020denoising}, which define a Markov chain that incrementally perturbs data $x_0 \in \mathbb{R}^d$ into Gaussian noise $x_T \sim \mathcal{N}(0,I)$ through intermediate latent variables {$x_1, \ldots, x_T$} defined as:
$q(x_t \mid x_{t-1}) = \mathcal{N}(x_t;\sqrt{1-\beta_t}\,x_{t-1},\,\beta_t I),$ where $\beta_t$ is a step size. A neural network models the reverse denoising step by predicting the noise component $\epsilon_\theta(x_t,t)$, thereby reconstructing $x_{t-1}$ from $x_t$. Training employs the MSE objective:
\begin{equation}
    \mathcal{L}_t = \mathbb{E}_{x_0,\epsilon,t}\,\|\epsilon-\epsilon_\theta(x_t,t)\|_2^2.    
    \label{eq:ddpm_mse_loss}
\end{equation}

\begin{wrapfigure}{r}{0.25\textwidth}
  \vspace{-0.4cm}
  \centering
  \includegraphics[width=\linewidth]{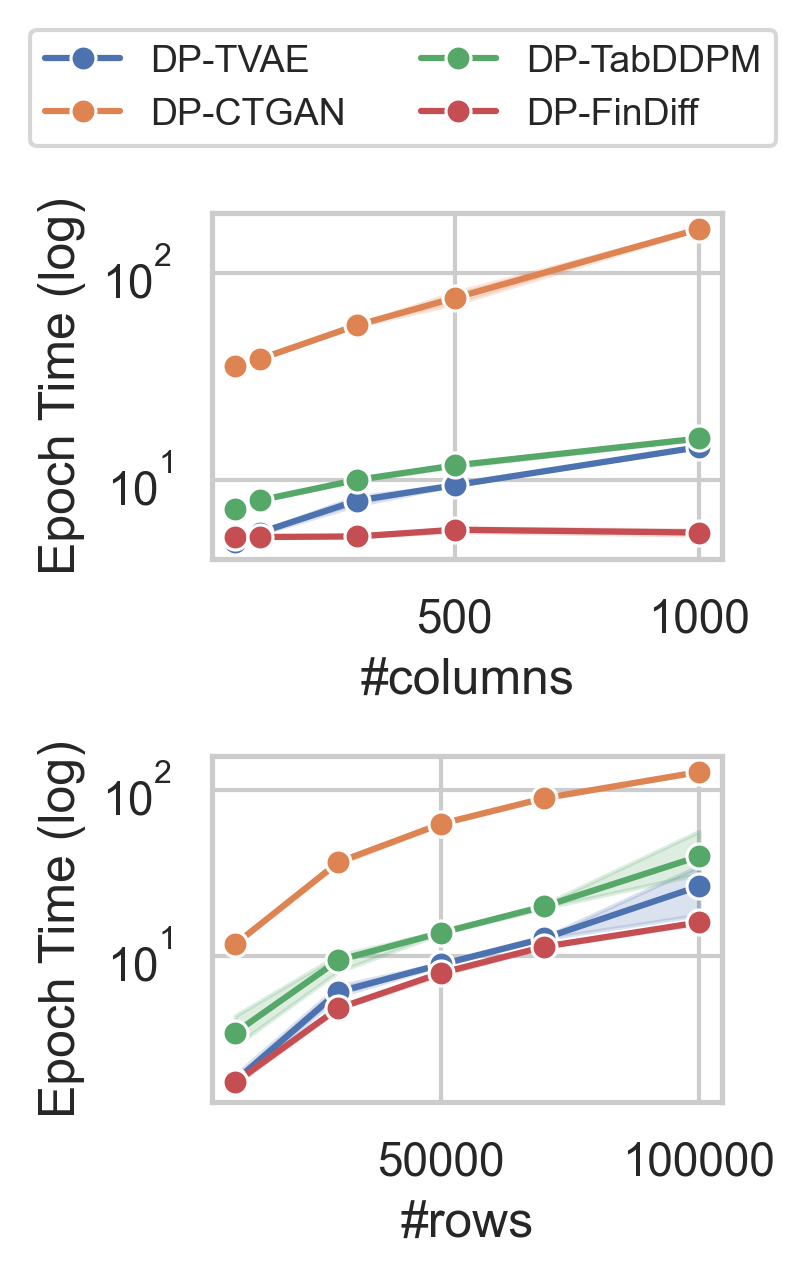}
  \caption{Training time per epoch as dataset size grows (rows \& columns).}
   \label{fig:speed_rows}
   \vspace{-0.8cm}
\end{wrapfigure}

\noindent\textbf{Differential privacy.}  
We employ \((\varepsilon,\delta)\)-DP~\cite{dwork2014algorithmic} enforced by DP-SGD~\cite{abadi2016deep} mechanism. For batch $B$, per-example gradients $g_{i}$ are clipped to norm $C$, averaged and Gaussian noise is added, yielding $\widehat{g}$:
\[
\widetilde{g}_{i} = g_i \cdot \min\!\Bigl(1, \tfrac{C}{\|g_i\|_2}\Bigr),
\quad \widehat{g} = \tfrac{1}{|B|}\sum_{i\in B} \widetilde{g}_{i} + \mathcal{N}(0,\sigma^2 I),
\]
before the model update. Privacy guaranties are accounted via PRV~\cite{gopi2021numerical}.  

\begin{figure*}[t!]
    \centering
    \includegraphics[width=0.9\linewidth]{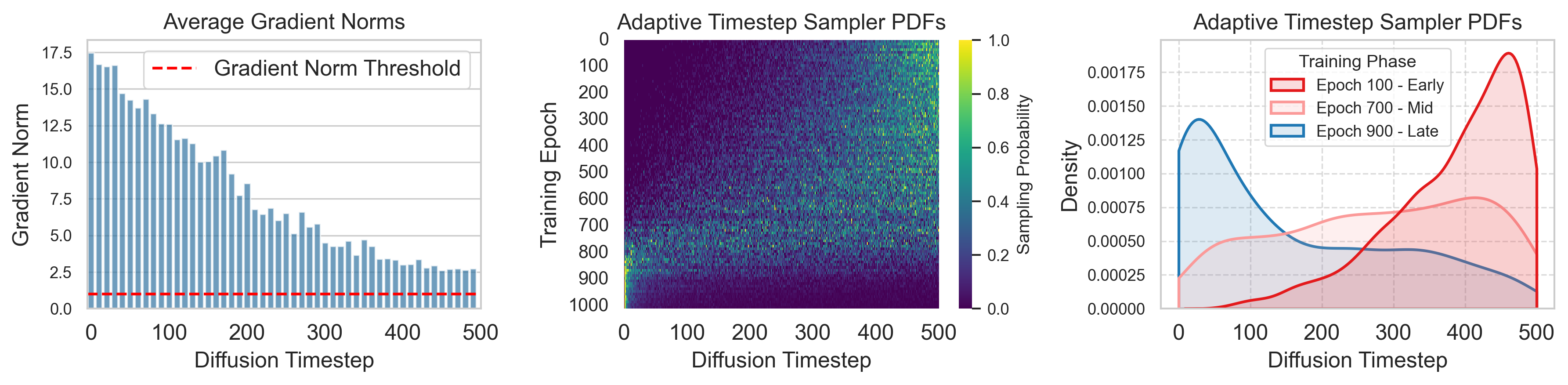}
    \includegraphics[width=0.08\linewidth]{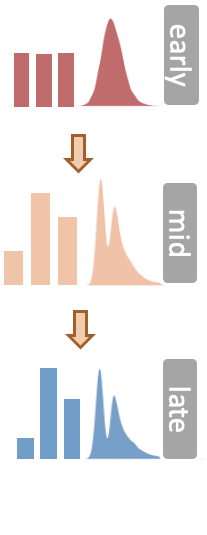}
    \caption{Timestep dynamics in \textit{DP-FinDiff}. (Left) Gradient norms with DP clipping threshold (red dashed). (Middle) AT sampler heatmap showing the shift from later to earlier diffusion timesteps over training. (Right) Sampling distributions at early, mid, and late phases ($\alpha_{\text{start}}=3$, $\alpha_{\text{end}}=-1$).}
    \label{fig:timestep_analysis}
    \vspace{-0.3cm}
\end{figure*}

\subsection{Model Adaptation for Tabular Data}

\noindent\textbf{Embedding space.}  
We adopt the core \textit{FinDiff}~\cite{sattarov2023findiff} design: each record $x_i=(x_i^{\text{num}},x_i^{\text{cat}})$ is mapped to a dense embedding: numerical features via $W x_i^{\text{num}}\in \mathbb{R}^{d_{\text{num}} \times d_e}$, categorical features via embeddings $E(x_i^{\text{cat}}) \in \mathbb{R}^{d_{\text{cat}} \times d_e}$. Concatenating produces unified embedding representaion:
\[
z_i^0 = [\,W x_i^{\text{num}} \;\|\; E(x_i^{\text{cat}})\,] \in \mathbb{R}^d, 
\quad d = (d_{\text{num}} + d_{\text{cat}}) \cdot d_e.
\]
The diffusion forward step perturbs $z_i^0$ using noise scheduler $\alpha_t$,
\[
z_i^t = \sqrt{\alpha_t}\,z_i^0 + \sqrt{1-\alpha_t}\,\epsilon,\quad \epsilon\sim\mathcal{N}(0,I),
\]
while the denoiser predicts $\epsilon_\theta(z_i^t,t)$, operating entirely in the embedding space. The reverse process inverts the embedding to data space projecting $z_i^t$ through linear heads for numerical and categorical features. Finally categorical codes are decoded via nearest-neighbor lookup in embedding space (L2). 

\noindent\textbf{Architecture.}  
The denoiser is a lightweight MLP with timestep embeddings. Dense embeddings eliminate one-hot sparsity, improving scalability and training efficiency. Fig.~\ref{fig:speed_rows} shows that \textit{DP-FinDiff} maintains significantly lower per-epoch training times than baselines with increasing rows or columns.

\subsection{DP-Aware Training Enhancements}
\label{subsec:theory-motivation}

In diffusion models, the reverse process reconstructs data by integrating the learned score across timesteps, making it sensitive to errors that accumulate throughout generation. Under DP-SGD, gradient clipping and noise distort score estimation unevenly across timesteps and features. Our two enhancements mitigate this effect in complementary ways: \emph{Adaptive Timestep (AT)} sampling allocates privacy budget to high-signal timesteps, reducing temporal error accumulation, while \emph{Feature-Aggregated (FA)} loss stabilizes per-sample gradient norms, reducing clipping bias and balancing feature contributions. Together, these mechanisms reduce distortion in score estimation, leading to stronger utility on downstream tasks and higher sample fidelity.

\vspace{-0.1cm}

\noindent\textbf{Adaptive Timestep (AT) Sampling.}  
In DDPMs, gradient norms decay with $t$ (Fig.~\ref{fig:timestep_analysis} left), so uniform sampling wastes updates on weak-signal steps. The training objective under uniform timestep sampling is:
$
L(\theta)=\tfrac{1}{T}\sum_{t=1}^T \mathbb{E}_i[\ell_{i,t}(\theta)].
$
Let the DP-effective signal at timestep $t$ be:
\[
s(t) := \mathbb{E}\bigl[\|\widetilde{g}_{i,t}\|_2\bigr], \quad 0 \le s(t)\le C.
\]
Uniform sampling spends updates on steps where $s(t)$ is small (late/noisy timesteps). By importance-sampling analysis, the variance-minimizing distribution is $q^\star_{\mathrm{DP}}(t)\propto s(t)$. Since estimating $s(t)$ online is noisy, we approximate it with
\[
P_k(t) = \tfrac{t^{\alpha_k}}{\sum_{s=1}^T s^{\alpha_k}},\qquad 
\alpha_k=\alpha_{\text{start}}+\tfrac{k}{K}(\alpha_{\text{end}}-\alpha_{\text{start}}).
\]
Transitioning $\alpha_k$ from positive to negative shifts probability mass from late timesteps to early ones, matching the observed dynamics in Fig.~\ref{fig:timestep_analysis} (mid \& right), the sampler heatmap shifts toward early steps and the induced distributions evolve smoothly across training.  
\emph{Intuition:} AT reallocates optimization effort across timesteps under a fixed privacy budget to timesteps with higher DP signal-to-noise ratio, without altering $(\varepsilon,\delta)$. Detailed derivation is described in~\ref{subsec:AT_sampling}.

\vspace{-0.1cm}

\textbf{DP-SGD clipping bias.} The clipped per-sample gradient at timestep $t$ is:
$
\widetilde{g}_{i,t} = g_{i,t}\cdot \min (1,\tfrac{C}{\|g_{i,t}\|_2}).
$
The expectation $\mathbb{E}[\widetilde{g}_{i,t}]$ differs from $g_{i,t}$ by a bias proportional to $\mathbb{E}[(\|g_{i,t}\|_2-C)_+]$. This bias becomes large when gradient norms have heavy tails.

\vspace{-0.1cm}

\begin{wrapfigure}{r}{0.45\textwidth}
  \centering
  \vspace{-0.3cm}
  \includegraphics[width=\linewidth]{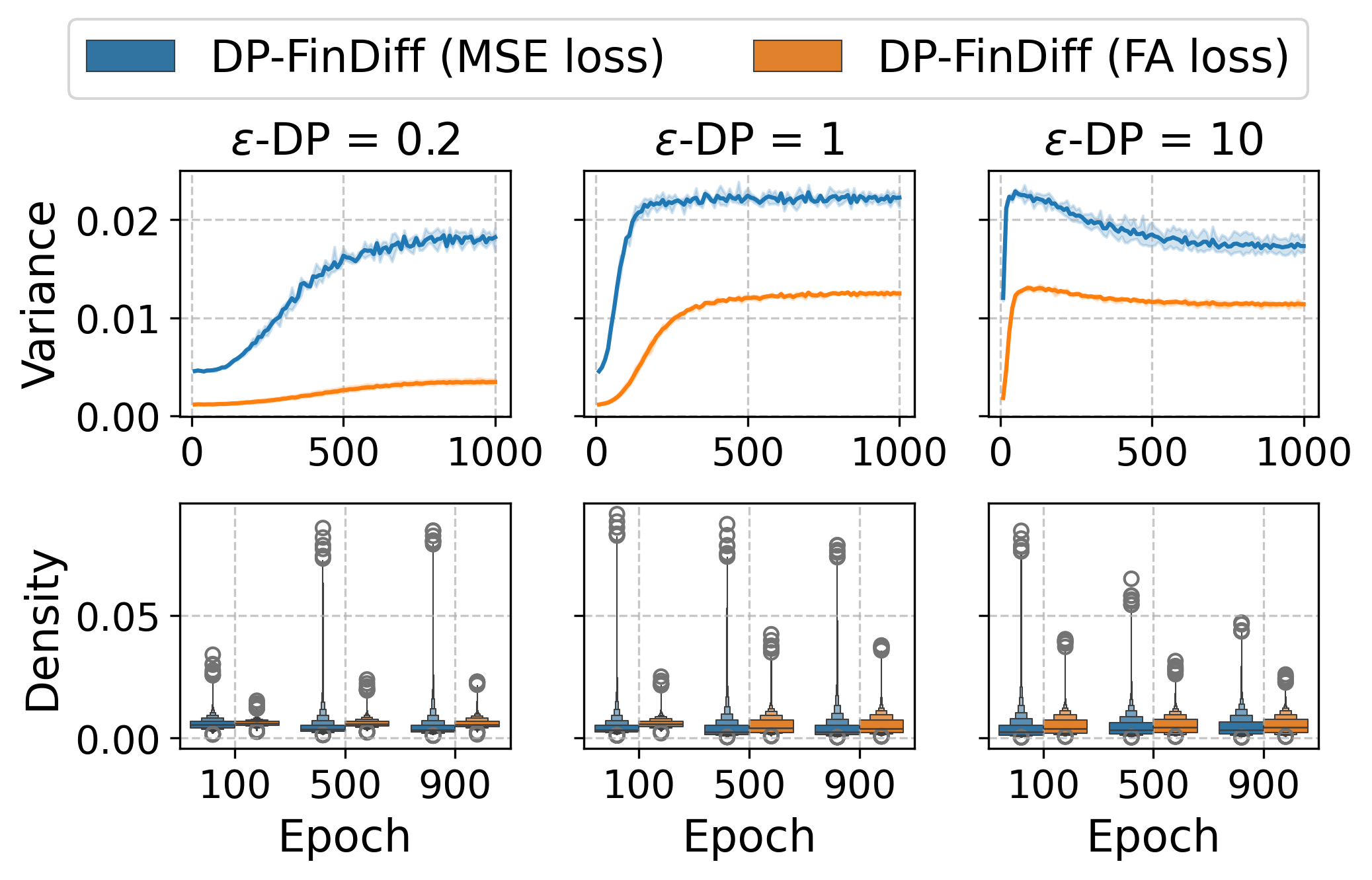}
    \caption{Per-sample gradient norms on Adult: MSE vs.\ FA loss. Top: normalized variance over epochs. Bottom: distributions of gradient norms at epochs 100/500/900, showing FA reduces skewness and variance.}
    \label{fig:grad_norms}  
    \vspace{-0.6cm}
\end{wrapfigure}

\textbf{Feature-Aggregated (FA) Loss.} The standard $\text{DDPM}$ objective (Eq. \ref{eq:ddpm_mse_loss}) calculates the per-sample loss by averaging reconstruction errors across $d$ features. For mixed-type tabular data, where standardized numeric features often retain unbounded non-Gaussian tails, this averaging mechanism leads to relative skewness of gradient norms with results in learning instability.

When calculating the per-sample $\mathbf{L_2}$ gradient norm $\|\mathbf{g}^{\mathrm{mse}}\|_2$ for clipping, the $\mathbf{1/d}$ scaling factor applied to every feature gradient component $\mathbf{g}_k$ disproportionately amplifies errors. Components below 1 vanish quadratically when squared, leaving the norm dominated by the few largest unscaled errors (e.g. outliers). This dynamics results in a highly skewed gradient norm distribution with heavy tails (Fig.~\ref{fig:grad_norms}), which raises clipping bias $\mathbf{E}[(\|\mathbf{g}\|_2-C)_+]$. To mitigate this, we define the Feature-Aggregated (FA) loss by summing the reconstruction errors without scaling:
\[
\mathcal{L}^{\mathrm{FA}}(\theta) \;=\;
\frac{1}{B} \sum_{i=1}^B \sum_{j=1}^d
\left( \epsilon_{i,j} - (\epsilon_\theta(z^{(i)}_{t_i}, t_i))_j \right)^2.
\]
The resulting gradient $\mathbf{g}^{\mathrm{FA}}$ uses the unscaled, natural magnitude of the feature errors, yielding a lighter tailed, less skewed distribution, thus efficiently preserving signal and enhancing utility without weakening privacy guaranties. Detailed derivation is described in~\ref{subsec:FA_loss}.


\noindent\textbf{Privacy guarantees.}  
Both AT and FA modify only the optimization objective, not the DP mechanism (clipping, noise, subsampling) or accounting. Thus the overall $(\varepsilon,\delta)$ privacy bound is preserved.

\begin{figure}[t!]
    \vspace{-0.2cm}
    \centering
    \begin{minipage}[b]{0.49\textwidth}
        \includegraphics[width=\linewidth]{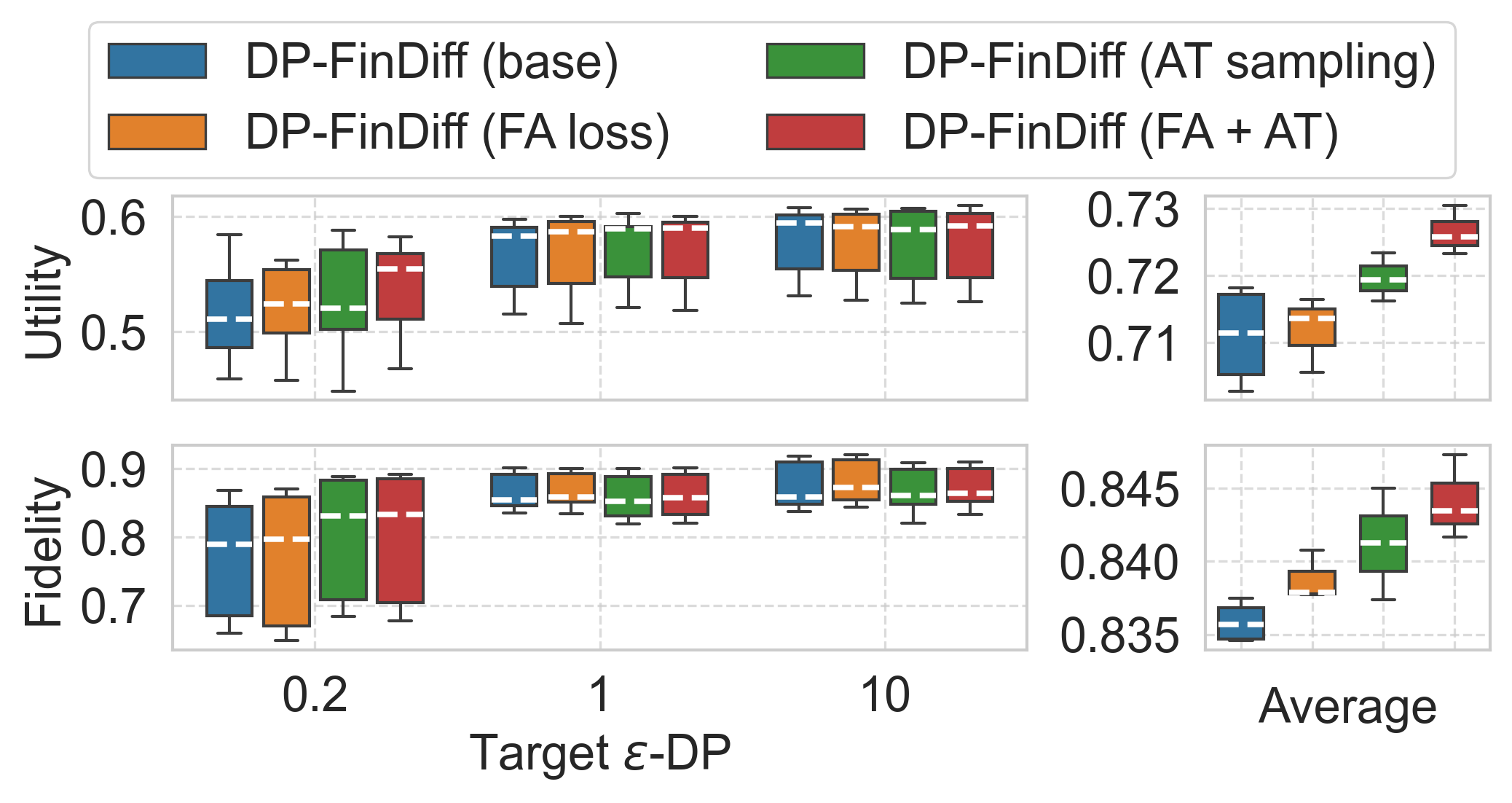}
        \caption{Utility and fidelity of \textit{DP-FinDiff} variants with \textit{FA}, \textit{AT}, and \textit{FA+AT}. Enhancements consistently boost results, most notably at $\varepsilon\!=\!0.2$.}
    \label{fig:design_improvements}        
    \end{minipage}
    \hfill
    \begin{minipage}[b]{0.49\textwidth}
        \includegraphics[width=\linewidth]{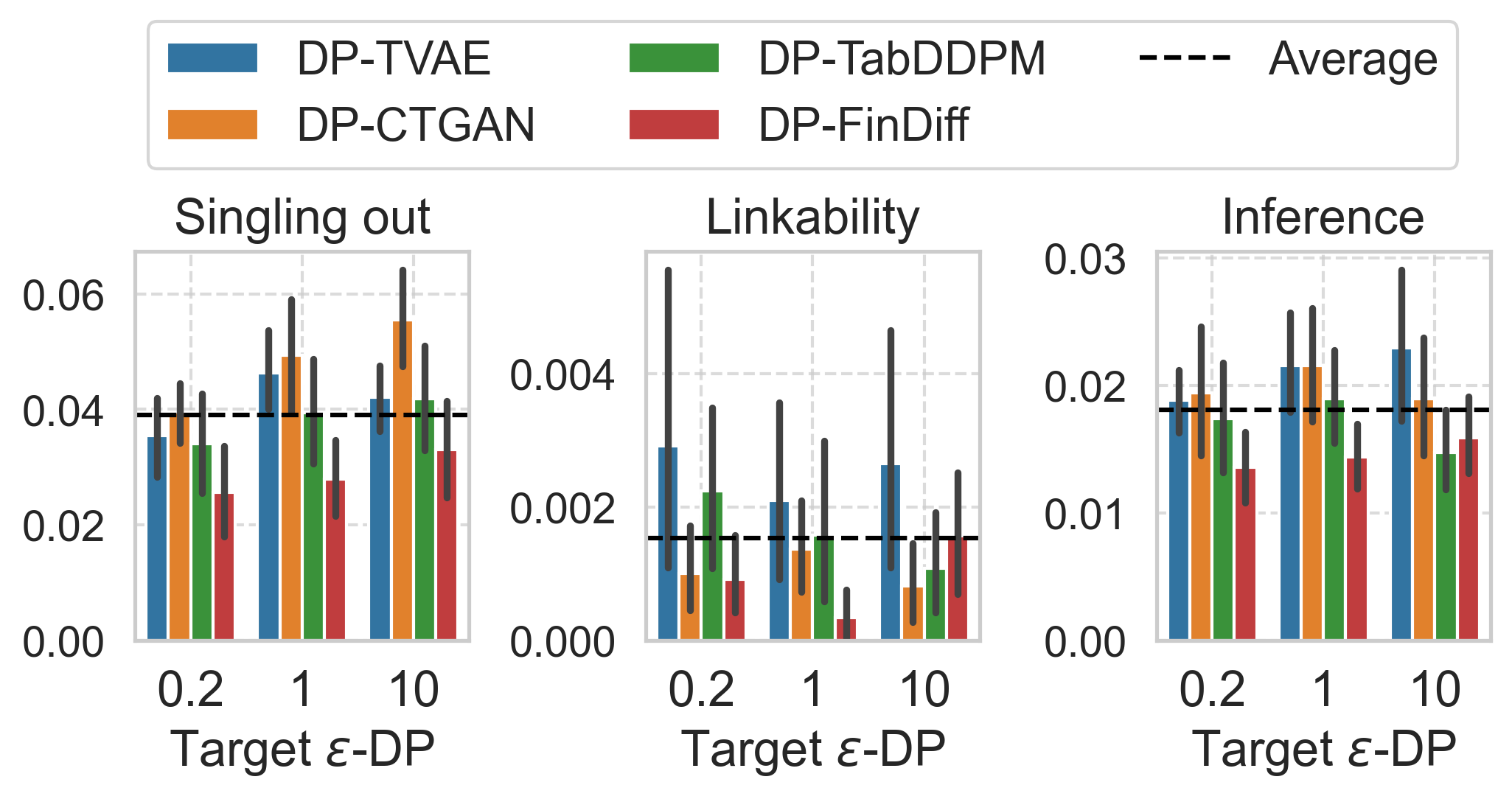}
        \caption{Privacy risk indicators across three attack types averaged over all datasets. Lower means safer; dashed lines show averages scores.}
    \label{fig:privacy_utility_scores}      
    \end{minipage}
    \vspace{-0.3cm}
\end{figure}

\section{Experimental results.}

\textbf{Setup.} We evaluate \textit{DP-FinDiff} on five mixed-type tabular datasets from finance and healthcare domains.
Models are trained under $(\varepsilon,\delta)$-DP using DP-SGD with Gaussian noise, gradient clipping threshold $C$=1.0, privacy budgets $\varepsilon\!\in\!\{0.2,1,10\}$ and $\delta\!=\!10^{-5}$. 
Baselines include DP-CTGAN, DP-TVAE, and DP-TabDDPM, re-implemented with identical DP accounting. 
Performance is assessed by: (i) downstream \textit{utility} via average ROC-AUC of five classifiers on synthetic-to-real transfer, (ii) \textit{fidelity} via column/row-wise KS/Wasserstein metrics, and (iii) \textit{privacy} risk via three types of simulated attacks. 
All results are averaged over five runs with fixed random seeds. Baseline methods were fairly compared using grid-searched architectures and identical training setups (\autoref{sec:experimental_setup}).

\textbf{Comparison Against Baselines.}
\textit{DP-FinDiff} is benchmarked against three DP models under $\varepsilon\!\in\!\{0.2,1,10\}$, with utility scores summarized in Table~\ref{tab:utility_scores} and privacy trends illustrated in Fig.~\ref{fig:privacy_utility_scores}.

\vspace{-0.1cm}

\textit{DP-FinDiff} achieves average highest utility scores across all datasets, outperforming \textit{DP-TVAE} by 16\%, \textit{DP-TabDDPM} by 33\%, and \textit{DP-CTGAN} by 42\% on average, with the largest gains at the strictest privacy level ($\varepsilon\!=\!0.2$).
Privacy risk metrics report the lowest attack success rates across all privacy budgets, confirming strong protection without compromising utility.

\vspace{-0.1cm}

\textit{DP-FinDiff} proves effective for privacy-preserving data synthesis with high analytical utility.

\noindent\textbf{Methodological Enhancements.}  
The impact of \textit{FA loss} and \textit{AT sampling} was evaluated through four setups: (i) base model, (ii) +FA, (iii) +AT, and (iv) +FA+AT, with results depicted on Fig.\ref{fig:design_improvements}.

\vspace{-0.1cm}

\textit{FA loss} and \textit{AT sampling} provide individual gains of 0.13\% and 1.22\%, jointly improving utility by 2.20\%, most pronounced under string privacy level ($\varepsilon\!=\!0.2$).
These enhancements also improve fidelity by 1\% in total, indicating complementary positive effects on structural preservation.

\vspace{-0.1cm}

These results underscore the value of specialized training strategies for DP diffusion models.

\noindent\textbf{Summary.} \textit{DP-FinDiff} surpasses DP baselines, and its enhancements further improve performance, validating its suitability for $\varepsilon$-DP tabular generation and downstream applications in sensitive domains.

\begin{table*}
  \centering
    \caption{Utility scores of \textit{DP-FinDiff} against baseline models and datasets at various $\varepsilon$-DP levels.}
    \label{tab:utility_scores}
    \fontsize{9.5pt}{6pt}\selectfont 
\begin{tabular}{clccccc}
  \toprule
    & & \multicolumn{5}{c}{\textbf{Dataset}} \\
    \cmidrule{3-7}
    \textbf{$\varepsilon$-DP} & \textbf{Model} & \textbf{Credit} & \textbf{Adult} & \textbf{Marketing} & \textbf{Payments} & \textbf{Diabetes} \\
    \midrule
 & DP-TVAE & 0.495±0.06 & 0.515±0.06 & 0.556±0.07 & 0.588±0.01 & 0.494±0.00 \\
 & DP-CTGAN & 0.495±0.03 & 0.513±0.07 & 0.517±0.09 & 0.490±0.01 & 0.501±0.00 \\
\multirow[t]{3}{*}{0.2} & DP-TabDDPM & 0.489±0.02 & 0.464±0.06 & 0.519±0.03 & 0.505±0.02 & 0.502±0.00 \\
 & DP-FinDiff & \textbf{0.583±0.05} & \textbf{0.708±0.02} & \textbf{0.757±0.02} & \textbf{0.737±0.03} & \textbf{0.572±0.01} \\
\cmidrule{1-7}
& DP-TVAE & 0.477±0.06 & 0.733±0.03 & 0.594±0.04 & 0.755±0.02 & 0.505±0.00 \\
 & DP-CTGAN & 0.486±0.03 & 0.515±0.06 & 0.458±0.13 & 0.499±0.02 & 0.493±0.01 \\
\multirow[t]{3}{*}{1} & DP-TabDDPM & 0.524±0.03 & 0.502±0.07 & 0.537±0.05 & 0.508±0.03 & 0.501±0.00 \\
 & DP-FinDiff & \textbf{0.674±0.03} & \textbf{0.768±0.03} & \textbf{0.797±0.01} & \textbf{0.770±0.01} & \textbf{0.584±0.01} \\
\cmidrule{1-7}
& DP-TVAE & 0.637±0.02 & \textbf{0.826±0.00} & 0.773±0.03 & \textbf{0.815±0.01} & 0.572±0.00 \\
 & DP-CTGAN & 0.487±0.05 & 0.542±0.07 & 0.513±0.09 & 0.537±0.08 & 0.497±0.01 \\
\multirow[t]{3}{*}{10} & DP-TabDDPM & 0.559±0.04 & 0.706±0.02 & 0.579±0.04 & 0.625±0.01 & 0.512±0.01 \\
 & DP-FinDiff & \textbf{0.697±0.01} & 0.792±0.01 & \textbf{0.804±0.02} & 0.799±0.01 & \textbf{0.582±0.00} \\
\bottomrule
\multicolumn{7}{l}{\scalebox{0.7}{*Scores are derived from the averaged results and standard deviations of five experiments, each initiated with distinct random seeds}}
\end{tabular}
\vspace{-0.3cm}
\end{table*}

\section{Limitations and Ethical Considerations}
\textit{DP-FinDiff} improves utility under strict privacy budgets, but its relative gains diminish at higher $\varepsilon$, suggesting room for better adaptation to lower noise regimes. Similarly, fairness across minority groups has not been systematically evaluated, and synthetic data may still propagate biases. We advise combining our method with bias audits and domain-specific supervision.

\section{Conclusion}
We introduced \textit{DP-FinDiff}, a diffusion-based framework for generating mixed-type tabular data under differential privacy. It addresses core DP modeling challenges: lowering encoding overhead, enhancing scalability, and mitigating utility loss through \textit{Adaptive Timestep sampling} and \textit{Feature-Aggregated loss}. To enable fair benchmarking, we implemented DP variants of TVAE, CTGAN, and TabDDPM. Experiments across financial and medical datasets show 16–42\% higher utility than DP baselines without compromising privacy, establishing \textit{DP-FinDiff} as a practical and effective solution for privacy-preserving data synthesis.

\begin{ack}
The authors thank the members of the Deutsche Bundesbank for
their valuable review and comments. The opinions expressed in this work are those of the authors and do not necessarily reflect the views of the Deutsche Bundesbank or the Swiss Federal Audit Office. 
\end{ack}

\bibliographystyle{plainnat}
\bibliography{references}

@inproceedings{sattarov2023findiff,
  title={Findiff: Diffusion models for financial tabular data generation},
  author={Sattarov, Timur and Schreyer, Marco and Borth, Damian},
  booktitle={Proceedings of the Fourth ACM International Conference on AI in Finance},
  pages={64--72},
  year={2023}
}

@inproceedings{kotelnikov2023tabddpm,
  title={{TabDDPM: Modelling Tabular Data with Diffusion Models}},
  author={Kotelnikov, Akim and Baranchuk, Dmitry and Rubachev, Ivan and Babenko, Artem},
  booktitle={International Conference on Machine Learning},
  pages={17564--17579},
  year={2023},
  organization={PMLR}
}

@article{cao2022survey,
  title={{A Survey on Generative Diffusion Model}},
  author={Cao, Hanqun and Tan, Cheng and Gao, Zhangyang and Xu, Yilun and Chen, Guangyong and Heng, Pheng-Ann and Li, Stan Z},
  journal={arXiv preprint arXiv:2209.02646},
  year={2022}
}

@inproceedings{schreyer2022federated,
  title={{Federated and Privacy-Preserving Learning of Accounting Data in Financial Statement Audits}},
  author={Schreyer, Marco and Sattarov, Timur and Borth, Damian},
  booktitle={Proceedings of the Third ACM International Conference on AI in Finance},
  pages={105--113},
  year={2022}
}

@inproceedings{zhao2021ctab,
  title={{CTAB-GAN: Effective Table Data Synthesizing}},
  author={Zhao, Zilong and Kunar, Aditya and Birke, Robert and Chen, Lydia Y},
  booktitle={Asian Conference on Machine Learning},
  year={2021},
}

@article{engelmann2021conditional,
  title={{Conditional Wasserstein GAN-based oversampling of tabular data for imbalanced learning}},
  author={Engelmann, Justin and Lessmann, Stefan},
  journal={Expert Systems with Applications},
  volume={174},
  pages={114582},
  year={2021},
  publisher={Elsevier}
}

@article{ho2020denoising,
  title={{Denoising Diffusion Probabilistic Models}},
  author={Ho, Jonathan and Jain, Ajay and Abbeel, Pieter},
  journal={Advances in Neural Information Processing Systems},
  volume={33},
  pages={6840--6851},
  year={2020}
}

@inproceedings{jordon2018pate,
  title={{PATE-GAN: Generating Synthetic Data with Differential Privacy Guarantees}},
  author={Jordon, James and Yoon, Jinsung and Van Der Schaar, Mihaela},
  booktitle={ICLR},
  year={2018}
}

@article{salem2018,
  title={{ML-Leaks: Model and Data Independent Membership Inference Attacks and Defenses on Machine Learning Models}},
  author={Salem, Ahmed and Zhang, Yang and Humbert, Mathias and Berrang, Pascal and Fritz, Mario and Backes, Michael},
  journal={arXiv preprint arXiv:1806.01246},
  year={2018}
}

@inproceedings{sohl2015deep,
  title={{Deep Unsupervised Learning Using Nonequilibrium Thermodynamics}},
  author={Sohl-Dickstein, Jascha and Weiss, Eric and Maheswaranathan, Niru and Ganguli, Surya},
  booktitle={International conference on machine learning},
  pages={2256--2265},
  year={2015},
  organization={PMLR}
}

@article{kingma2014adam,
  title={Adam: A method for stochastic optimization},
  author={Kingma, Diederik P and Ba, Jimmy},
  journal={arXiv preprint arXiv:1412.6980},
  year={2014}
}

@article{pytorch,
  title={Pytorch: An imperative style, high-performance deep learning library},
  author={Paszke, Adam and Gross, Sam and Massa, Francisco and Lerer, Adam and Bradbury, James and Chanan, Gregory and Killeen, Trevor and Lin, Zeming and Gimelshein, Natalia and others},
  journal={NeurIPS},
  year={2019}
}

@article{tvae_ctgan,
  title={Modeling tabular data using conditional gan},
  author={Xu, Lei and Skoularidou, Maria and Cuesta-Infante, Alfredo and Veeramachaneni, Kalyan},
  journal={NeurIPS},
  volume={32},
  year={2019}
}

@article{dockhorn2022differentially,
  title={Differentially private diffusion models},
  author={Dockhorn, Tim and Cao, Tianshi and Vahdat, Arash and Kreis, Karsten},
  journal={arXiv preprint arXiv:2210.09929},
  year={2022}
}

@article{opacus,
  title={Opacus: {U}ser-Friendly Differential Privacy Library in {PyTorch}},
  author={Ashkan Yousefpour and Igor Shilov and Alexandre Sablayrolles and Davide Testuggine and Karthik Prasad and Mani Malek and John Nguyen and Sayan Ghosh and Akash Bharadwaj and Jessica Zhao and Graham Cormode and Ilya Mironov},
  journal={arXiv preprint arXiv:2109.12298},
  year={2021}
}

@misc{anonymeter,
  doi = {https://doi.org/10.56553/popets-2023-0055},
  url = {https://petsymposium.org/popets/2023/popets-2023-0055.php},
  journal = {Proceedings of Privacy Enhancing Technologies Symposium},
  year = {2023},
  author = {Giomi, Matteo and Boenisch, Franziska and Wehmeyer, Christoph and Tasnádi, Borbála},
  title = {A Unified Framework for Quantifying Privacy Risk in Synthetic Data},
}

@inproceedings{fan2020survey,
  title={A survey of differentially private generative adversarial networks},
  author={Fan, Liyue},
  booktitle={The AAAI Workshop on Privacy-Preserving Artificial Intelligence},
  volume={8},
  year={2020}
}

@inproceedings{dwork2006our,
  title={Our data, ourselves: Privacy via distributed noise generation},
  author={Dwork, Cynthia and Kenthapadi, Krishnaram and McSherry, Frank and Mironov, Ilya and Naor, Moni},
  booktitle={Advances in Cryptology-EUROCRYPT 2006. Proceedings 25},
  pages={486--503},
  year={2006},
  organization={Springer}
}

@inproceedings{abadi2016deep,
  title={Deep learning with differential privacy},
  author={Abadi, Martin and Chu, Andy and Goodfellow, Ian and McMahan, H Brendan and Mironov, Ilya and Talwar, Kunal and Zhang, Li},
  booktitle={Proceedings of the 2016 ACM SIGSAC on computer and communications security},
  year={2016}
}

@article{zheng2020pcal,
  title={Pcal: A privacy-preserving intelligent credit risk modeling framework based on adversarial learning},
  author={Zheng, Yuli and Wu, Zhenyu and Yuan, Ye and Chen, Tianlong and Wang, Zhangyang},
  journal={arXiv preprint arXiv:2010.02529},
  year={2020}
}

@inproceedings{byrd2020differentially,
  title={Differentially private secure multi-party computation for federated learning in financial applications},
  author={Byrd, David and Polychroniadou, Antigoni},
  booktitle={Proceedings of the First ACM International Conference on AI in Finance},
  pages={1--9},
  year={2020}
}

@article{dwork2014algorithmic,
  title={The algorithmic foundations of differential privacy},
  author={Dwork, Cynthia and Roth, Aaron and others},
  journal={Foundations and Trends{\textregistered} in Theoretical Computer Science},
  year={2014},
  publisher={Now Publishers, Inc.}
}

@software{Zychlinski_dython_2018,
  author = {Zychlinski, Shaked},
  title = {{dython}},
  year = {2018},
  url = {https://github.com/shakedzy/dython},
  doi = {10.5281/zenodo.12698421}
}

@inproceedings{lin2021privacy,
  title={On the privacy properties of gan-generated samples},
  author={Lin, Zinan and Sekar, Vyas and Fanti, Giulia},
  booktitle={International Conference on Artificial Intelligence and Statistics},
  pages={1522--1530},
  year={2021},
  organization={PMLR}
}

@article{lin2019generating,
  title={Generating high-fidelity, synthetic time series datasets with doppelganger},
  author={Lin, Zinan and Jain, Alankar and Wang, Chen and Fanti, Giulia and Sekar, Vyas},
  journal={arXiv preprint arXiv:1909.13403},
  year={2019}
}

@inproceedings{lee2023codi,
  title={Codi: Co-evolving contrastive diffusion models for mixed-type tabular synthesis},
  author={Lee, Chaejeong and Kim, Jayoung and Park, Noseong},
  booktitle={International Conference on Machine Learning},
  pages={18940--18956},
  year={2023},
  organization={PMLR}
}

@article{zhang2023mixed,
  title={Mixed-type tabular data synthesis with score-based diffusion in latent space},
  author={Zhang, Hengrui and Zhang, Jiani and Srinivasan, Balasubramaniam and Shen, Zhengyuan and Qin, Xiao and Faloutsos, Christos and Rangwala, Huzefa and Karypis, George},
  journal={arXiv preprint arXiv:2310.09656},
  year={2023}
}

@article{suh2023autodiff,
  title={Autodiff: combining auto-encoder and diffusion model for tabular data synthesizing},
  author={Suh, Namjoon and Lin, Xiaofeng and Hsieh, Din-Yin and Honarkhah, Merhdad and Cheng, Guang},
  journal={arXiv preprint arXiv:2310.15479},
  year={2023}
}

@inproceedings{fang2022dp,
  title={Dp-ctgan: Differentially private medical data generation using ctgans},
  author={Fang, Mei Ling and Dhami, Devendra Singh and Kersting, Kristian},
  booktitle={International Conference on Artificial Intelligence in Medicine},
  pages={178--188},
  year={2022},
  organization={Springer}
}

@article{kunar2021dtgan,
  title={DTGAN: Differential private training for tabular GANs},
  author={Kunar, Aditya and Birke, Robert and Zhao, Zilong and Chen, Lydia},
  journal={arXiv preprint arXiv:2107.02521},
  year={2021}
}

@article{truda2023generating,
  title={Generating tabular datasets under differential privacy},
  author={Truda, Gianluca},
  journal={arXiv preprint arXiv:2308.14784},
  year={2023}
}

@inproceedings{guan2024generating,
  title={Generating Privacy-preserving Educational Data Records with Diffusion Model},
  author={Guan, Quanlong and Yu, Yanchong and Huang, Xiujie and Fang, Liangda and He, Chaobo and Wu, Lusheng and Luo, Weiqi and Chen, Guanliang},
  booktitle={Companion Proceedings of the ACM on Web Conference 2024},
  pages={806--809},
  year={2024}
}

@article{fonseca2023tabular,
  title={Tabular and latent space synthetic data generation: a literature review},
  author={Fonseca, Joao and Bacao, Fernando},
  journal={Journal of Big Data},
  volume={10},
  number={1},
  pages={115},
  year={2023},
  publisher={Springer}
}

@article{li2025diffusion,
  title={Diffusion Models for Tabular Data: Challenges, Current Progress, and Future Directions},
  author={Li, Zhong and Huang, Qi and Yang, Lincen and Shi, Jiayang and Yang, Zhao and van Stein, Niki and B{\"a}ck, Thomas and van Leeuwen, Matthijs},
  journal={arXiv:2502.17119},
  year={2025}
}

@inproceedings{liu2024entity,
  title={Entity-based Financial Tabular Data Synthesis with Diffusion Models},
  author={Liu, Changshuo and Liu, Canyao},
  booktitle={Proceedings of the 5th ACM ICAIF},
  year={2024}
}

@inproceedings{schreyer2024imb,
  title={Imb-FinDiff: Conditional Diffusion Models for Class Imbalance Synthesis of Financial Tabular Data},
  author={Schreyer, Marco and Sattarov, Timur and Sim, Alexander and Wu, Kesheng},
  booktitle={Proceedings of the 5th ACM ICAIF},
  pages={617--625},
  year={2024}
}

@article{hassan2023deep,
  title={Deep generative models, synthetic tabular data, and differential privacy: An overview and synthesis},
  author={Hassan, Conor and Salomone, Robert and Mengersen, Kerrie},
  journal={arXiv preprint arXiv:2307.15424},
  year={2023}
}

@article{kim2024generative,
  title={Generative models for tabular data: A review},
  author={Kim, Dong-Keon and Ryu, DongHeum and Lee, Yongbin and Choi, Dong-Hoon},
  journal={Journal of Mechanical Science and Technology},
  volume={38},
  number={9},
  pages={4989--5005},
  year={2024},
  publisher={Springer}
}

@article{gopi2021numerical,
  title={Numerical composition of differential privacy},
  author={Gopi, Sivakanth and Lee, Yin Tat and Wutschitz, Lukas},
  journal={NeurIPS},
  volume={34},
  pages={11631--11642},
  year={2021}
}

@article{liu2024generative,
  title={Generative AI model privacy: a survey},
  author={Liu, Yihao and Huang, Jinhe and Li, Yanjie and Wang, Dong and Xiao, Bin},
  journal={Artificial Intelligence Review},
  volume={58},
  number={1},
  pages={33},
  year={2024},
  publisher={Springer}
}

@inproceedings{shokri2017membership,
  title={Membership inference attacks against machine learning models},
  author={Shokri, Reza and Stronati, Marco and Song, Congzheng and Shmatikov, Vitaly},
  booktitle={2017 IEEE symposium on security and privacy (SP)},
  pages={3--18},
  year={2017},
  organization={IEEE}
}

@inproceedings{carlini2021extracting,
  title={Extracting training data from large language models},
  author={Carlini, Nicholas and Tramer, Florian and Wallace, Eric and Jagielski, Matthew and Herbert-Voss, Ariel and Lee, Katherine and Roberts, Adam and Brown, Tom and Song, Dawn and Erlingsson, Ulfar and others},
  booktitle={30th USENIX security symposium (USENIX Security 21)},
  pages={2633--2650},
  year={2021}
}

@inproceedings{carlini2023extracting,
  title={Extracting training data from diffusion models},
  author={Carlini, Nicolas and Hayes, Jamie and Nasr, Milad and Jagielski, Matthew and Sehwag, Vikash and Tramer, Florian and Balle, Borja and Ippolito, Daphne and Wallace, Eric},
  booktitle={32nd USENIX security symposium (USENIX Security 23)},
  pages={5253--5270},
  year={2023}
}


\newpage

\appendix

\section{Experimental Setup}
\label{sec:experimental_setup}

This section describes the details of the conducted experiments, encompassing used datasets, data preparation steps, model architecture including hyperparameters, and evaluation metrics.

\subsection{Datasets and Data Preparation}
\label{subsec:datasets}

In our experiments, we utilized the following five real-world and mixed-type tabular datasets (categorical and numerical) from finance and healthcare domains:

\begin{enumerate}

    \item \textbf{Credit Default}\footnote{\url{https://archive.ics.uci.edu/ml/datasets/default+of+credit+card+clients}} (Credit): This dataset includes customers default payments records (e.g., payment history and bill statements) from April to September 2005. 

    \item \textbf{Census Income Data}\footnote{\url{https://archive.ics.uci.edu/dataset/2/adult}} (Adult): This dataset contains demographic information from the 1994 U.S. Census to predict whether a person earns more than \$50,000 per year. 

    \item \textbf{Marketing Data}\footnote{\url{https://www.kaggle.com/datasets/prakharrathi25/banking-dataset-marketing-targets/data}} (Marketing): This dataset contains customer records of a bank from 2008 to 2010, used to predict whether a client will subscribe to a term deposit. 

    \item \textbf{City of Philadelphia}\footnote{\url{https://data.phila.gov/visualizations/payments}} (Payments): The dataset contains checks and direct deposit payments made by the City of Philadelphia during the 2017 fiscal year. 

    \item \textbf{Diabetes hospital data \footnote{\url{https://www.kaggle.com/datasets/brandao/diabetes}}} (Diabetes): The dataset contains clinical care records collected by 130 US hospitals between the years 1999-2008.

\end{enumerate}

\begin{table}[h!]
\centering
\caption{Descriptive statistics of datasets. 
Raw Dim shows the number of numerical and categorical features. 
Encoded Dim compares the dimensionality after embedding-based encoding (\textbf{ours}, $d_e\!=\!2$) 
and one-hot encoding (\textbf{baselines}).}
\label{tab:dataset_stats_final}
\fontsize{9pt}{10pt}\selectfont
\begin{tabular}{lcccc|cc}
\toprule
& & \multicolumn{2}{c}{\textbf{Raw Dim}} & &
\multicolumn{2}{c}{\textbf{Encoded Dim}} \\
\cmidrule(lr){3-4} \cmidrule(lr){6-7}
\textbf{Dataset} & \textbf{\#Rows} & \textbf{\#Num} & \textbf{\#Cat} & &
\textbf{\#Emb. Dim (Ours)} & \textbf{\#One-Hot Dim (Baselines)} \\
\midrule
Credit       & 30{,}000   & 13 & 9  && $(13{+}9)\!\times\!2 = 44$  & $13 + \sum_{j=1}^{9}|\mathcal{V}_j| = 144$ \\
Adult        & 32{,}561   & 3  & 10 && $(3{+}10)\!\times\!2 = 26$  & $3 + \sum_{j=1}^{10}|\mathcal{V}_j| = 274$ \\
Marketing    & 45{,}211   & 6  & 10 && $(6{+}10)\!\times\!2 = 32$  & $6 + \sum_{j=1}^{10}|\mathcal{V}_j| = 129$ \\
Payments$^\dagger$ & 100{,}000 & 1  & 10 && $(1{+}10)\!\times\!2 = 22$  & $1 + \sum_{j=1}^{10}|\mathcal{V}_j| = 2115$ \\
Diabetes     & 101{,}767  & 8  & 40 && $(8{+}40)\!\times\!2 = 96$  & $8 + \sum_{j=1}^{40}|\mathcal{V}_j| = 2438$ \\
\bottomrule
\multicolumn{7}{l}{\footnotesize $\mathcal{V}_j$ denotes the set of unique categories for the $j$-th categorical feature.} \\
\multicolumn{7}{l}{\footnotesize $^\dagger$Subsampled from 238{,}894 due to memory constraints for baseline models.}
\end{tabular}
\end{table}

 To standardize the numeric attributes, we employed standard scaler, as implemented in the scikit-learn library.~\footnote{\url{https://scikit-learn.org/stable/modules/generated/sklearn.preprocessing.StandardScaler.html}} For the categorical attributes, we utilized embedding techniques following the approach outlined in~\cite{sattarov2023findiff}. Due to memory limitations faced during the training of baseline models the Payment dataset had to be reduced to 100,000 observations of the total 238,894.

\subsection{Model Architecture and Hyperparameters}

In the following, we describe the architectural choices and hyperparameter settings used for optimizing the \textit{DP-FinDiff} model. For every model training is performed for up to \scalebox{0.95}{$K=1,000$} epochs with a mini-batch size of 128, using the Adam optimizer~\cite{kingma2014adam} configured with parameters \scalebox{0.95}{$\beta_{1}=0.9$} and \scalebox{0.95}{$\beta_{2}=0.999$}.

\noindent \textbf{Diffusion Model.\footnote{All model training was conducted using PyTorch v2.2.1~\cite{pytorch}.}} 
The neural network architecture consists of two fully connected layers with 512 hidden units each. The diffusion process follows the settings of \textit{FinDiff}~\cite{sattarov2023findiff}, employing \scalebox{0.95}{$T=500$} diffusion steps and a linear noise scheduler with \scalebox{0.95}{$\beta_{\text{start}}=0.0001$} and \scalebox{0.95}{$\beta_{\text{end}}=0.02$}. Categorical attributes are represented using 2-dimensional embeddings.  

\noindent \textbf{Differential Privacy.\footnote{DP training and accounting were performed using Opacus v1.4.1~\cite{opacus}.}}  
We adopt the privacy settings from Dockhorn et al.~\cite{dockhorn2022differentially} using $\varepsilon\in\{0.2, 1, 10\}$ to reflect high, medium, and low privacy levels, respectively. Gradient clipping is applied with a per-sample norm of \scalebox{0.95}{$C=1.0$}. The probability of information leakage is bounded by \scalebox{0.95}{$\delta = 10^{-5}$}.

\noindent \textbf{Baselines.} We benchmark our model against DP versions of \textit{TVAE}~\cite{tvae_ctgan}, \textit{CTGAN}~\cite{tvae_ctgan}, and \textit{TabDDPM}~\cite{kotelnikov2023tabddpm}, using the best configurations obtained from a grid hyperparameter search to ensure fair evaluation (Table \ref{tab:dp_baseline_grid}).

\begin{table*}[t]
\centering
\caption{Grid search ranges for architecture and training hyperparameters across all DP baselines. DP accounting (batch size, sampling rate, and clip norm) was matched per privacy budget $\varepsilon\!\in\!\{0.2,1,10\}$.}
\label{tab:dp_baseline_grid}
\resizebox{\textwidth}{!}{
\begin{tabular}{lccc}
\toprule
\textbf{Hyperparameter} & \textbf{DP-TVAE} & \textbf{DP-CTGAN} & \textbf{DP-TabDDPM} \\
\midrule
Latent dimension / width & \{16, 32, 64, 128\} & – & \{128, 256, 512\} \\
Hidden layers (encoder/decoder) & \{[128,128], [256,256], [512,512]\} & \{[128,128], [256,256], [512,512]\} & \{[128,128], [256,256], [512,512]\} \\
Activation & \{ReLU, LeakyReLU(0.2)\} & \{ReLU, LeakyReLU(0.2)\} & \{ReLU, LeakyReLU(0.2), SiLU\} \\
Learning rate (Adam) & \{1e{-}3, 1e{-}4, 1e{-}5\} & \{1e{-}3, 1e{-}4, 1e{-}5\} & \{1e{-}3, 1e{-}4, 1e{-}5\} \\

Diffusion steps $T$ & – & – & \{200, 500, 1000\} \\
Diffusion scheduler & – & – & \{linear, cosine\} \\
Timestep embedding dim & – & – & \{32, 64, 128\} \\

\bottomrule
\end{tabular}
}
\end{table*}

\subsection{Compute Resources and Runtime}
\label{app:compute}

\textbf{Hardware and Software.}
All experiments were executed on a single \textbf{NVIDIA DGX-1} workstation equipped with
8× \texttt{NVIDIA V100 (32 GB)} GPUs, dual \texttt{Intel Xeon E5-2698 v4} CPUs (40 cores total),
and 512 GB RAM, running \texttt{Ubuntu 22.04}, CUDA 11.8, and cuDNN 8.9.
Training used \texttt{PyTorch 2.2.1} and \texttt{Opacus 1.4.1}; all other dependencies are listed in
the released code. Each experiment was executed on a \emph{single} GPU.
We fixed random seeds $\{0,1,2,3,4\}$ and enabled deterministic cuDNN for reproducibility.

\textbf{Runtime Summary.}
Table~\ref{tab:runtime} reports the average \emph{total wall-clock time} (in GPU-hours) required to complete single run of 1000 training epochs for each model–dataset combination. 

\begin{table}[h!]
\centering
\caption{Total runtime per model and dataset (GPU-hours, averaged over five seeds).}
\label{tab:runtime}
\begin{tabular}{lccccc}
\toprule
\textbf{Model} & \textbf{Credit} & \textbf{Adult} & \textbf{Marketing} & \textbf{Payments} & \textbf{Diabetes} \\
\midrule
DP-TVAE   & 2.30±0.06 & 3.5.8±0.06 & 2.60±0.09 & 6.11±0.06 & 6.01±0.01 \\
DP-CTGAN    & 4.50±0.11 & 5.20±0.11 & 4.41±0.07 & 40.29±0.05 & 44.03±0.06 \\
DP-TabDDPM & 3.70±0.02 & 4.40±0.87 & 3.50±0.01 & 8.22±0.02 & 9.08±0.19 \\
DP-FinDiff (ours) & \textbf{1.09±0.06} & \textbf{1.47±0.06} & \textbf{1.10±0.03} & \textbf{3.51±0.01} & \textbf{3.65±0.06} \\
\bottomrule
\end{tabular}
\end{table}

\subsection{Evaluation Metrics}

A comprehensive set of standard evaluation metrics, including \textbf{privacy}, \textbf{utility}, and \textbf{fidelity}, is employed to evaluate the model's effectiveness. 
These metrics represent diverse aspects of data generation quality, providing a broad view of model performance.

\noindent \textbf{Privacy.}\footnote{The estimation of privacy risks was conducted using the \textit{Anonymeter} library~\cite{anonymeter}.} 
The privacy metric evaluates the extent to which synthetic data prevents re-identification of individuals in the original dataset. In this study, privacy is assessed through three core indicators of factual anonymization, aligned with GDPR guidelines.~\footnote{\url{https://ec.europa.eu/justice/article-29/documentation/opinion-recommendation/files/2014/wp216_en.pdf}} Specifically, we adopt evaluators to quantify the risks of (i) \textit{singling out}, (ii) \textit{linkability}, and (iii) \textit{inference}, which represent key attack vectors synthetic data should resist. With \(\bm{S}\) as synthetic dataset and \(\bm{X}\) as the real dataset, the risks are defined as:

\begin{itemize}

\item \textit{Singling Out Risk (SOR):} Measures the probability \(SOR(X, S)\) that a synthetic record corresponds uniquely to a real individual. This reflects the risk of identifying someone in the original dataset based on a distinct synthetic entry.  

\item \textit{Linkability Risk (LR):} Quantifies the proportion \(LR(X, S)\) of successful matches between synthetic and real records by linking shared attributes. This captures the potential for re-identification through a partial attributes matching.  

\item \textit{Inference Risk (IR):} Evaluates the ability \(IR(X, S)\) of an attacker to predict a hidden (secret) attribute from auxiliary data. This measures how much sensitive information could be inferred by exploiting relationships in the synthetic dataset.  

\end{itemize}

\noindent This empirical, attack-based evaluation effectively reflects real-world privacy risks and offers a practical measure of privacy. 

\noindent \textbf{Utility.} The utility of synthetic data reflects its efficacy in supporting downstream machine learning tasks and its functional equivalence to real-world data. We quantify utility by training classifiers on synthetic datasets (\(S_{\text{train}}\)) and evaluating their performance on the original test set (\(X_{\text{test}}\)). This approach assesses how well the synthetic data captures the statistical properties needed for accurate model training. The overall utility score \(\Phi\) is computed as the average accuracy across all classifiers, formalized as:  

\begin{equation}
\Phi = \frac{1}{N} \sum_{i=1}^{N} \Theta_i(S_{\text{train}}, X_{\text{test}}),
\end{equation}

\noindent where \(\Theta_i\) is the \textit{Area Under the Receiver Operating Characteristic Curve} (\textit{ROC AUC})\footnote{\url{https://scikit-learn.org/stable/modules/generated/sklearn.metrics.roc_auc_score.html}} of the \(i\)-th classifier. To ensure a robust evaluation, we use \(N=5\) classifiers: \textit{Random Forest}, \textit{Decision Tree}, \textit{Logistic Regression}, \textit{AdaBoost}, and \textit{MLP Classifier}.

\noindent \textbf{Fidelity.}\footnote{The row fidelity computation were performed using the Dython library v0.7.5~\cite{Zychlinski_dython_2018}.} Fidelity evaluates how closely synthetic data replicates the statistical properties of real data, considering both column-level and row-level comparisons.  

For column fidelity, the similarity between corresponding attributes in the real (\(X\)) and synthetic (\(S\)) datasets is measured. Numeric attributes use the \textit{Wasserstein similarity} \(WS(x^d, s^d)\) to quantify distributional differences, while categorical attributes are evaluated using the \textit{Jensen-Shannon divergence} \(JS(x^d, s^d)\). These measures are combined into the column fidelity score \(\Omega_{col}\):  


\begin{equation}
\Omega_{col} =
\begin{cases}
  1 - WS(x^d, s^d), & d \in \mathcal{D}_{\text{num}}, \\
  1 - JS(x^d, s^d), & d \in \mathcal{D}_{\text{cat}}.
\end{cases}
\end{equation}

\noindent The overall column fidelity is the mean \(\Omega_{col}\) over all attributes in \(S\).  

Row fidelity assesses structural relationships between attribute pairs. For numeric pairs, the discrepancy between real and synthetic \textit{Pearson correlations} is computed as \(PC(x^{a,b}, s^{a,b}) = |\rho(x^a, x^b) - \rho(s^a, s^b)|\). For categorical pairs, the \textit{Theil’s U} coefficient \(TU(x^{a,b}, s^{a,b})\) quantifies differences in associations. These metrics are combined into the row fidelity score \(\Omega_{row}\):  


\begin{equation}
    \Omega_{row}=
    \begin{cases}
        1-PC(x^{a,b}, s^{a,b}), & d \in \mathcal{D}_{\text{num}}, \\
        1-TU(x^{a,b}, s^{a,b}), & d \in \mathcal{D}_{\text{cat}}
    \end{cases}
\end{equation}

\noindent The total row fidelity is calculated as the mean of \(\Omega_{row}\) over all attribute pairs. Finally, the aggregate fidelity score \(\Omega(X, S)\) is defined as the average of column and row fidelity scores, providing a holistic measure of similarity between real and synthetic datasets.  

\section{Detailed Theoretical Motivation}
\label{app:theory}

\paragraph{DP-SGD recap.}
For per-example gradient $g_{i,t}=\nabla_\theta \ell_{i,t}(\theta)$, DP-SGD clips and perturbs:
\[
\widetilde{g}_{i,t} = g_{i,t}\cdot \min\!\left(1,\tfrac{C}{\|g_{i,t}\|_2}\right), 
\qquad 
\widehat{g}_t = \tfrac{1}{B}\sum_{i\in B} \widetilde{g}_{i,t} + \mathcal{N}(0,\sigma^2 I).
\]
The bias term is
\[
\mathbb{E}[\widetilde{g}_{i,t}] 
= \mathbb{E}[g_{i,t}] - \mathbb{E}\!\left[\tfrac{(\|g_{i,t}\|_2-C)_+}{\|g_{i,t}\|_2}\,g_{i,t}\right],
\]
so large clipping bias arises when $\|g_{i,t}\|_2$ has heavy tails.

\subsection{Feature-Aggregated (FA) Loss.}
\label{subsec:FA_loss}
The empirical observation (Figure 4) that the Feature-Aggregated loss ($\mathcal{L}^{\mathrm{FA}}$) results in a distribution of gradient norms with significantly lower variance and skewness compared to the Mean-aggregated loss ($\mathcal{L}^{\mathrm{mean}}$) originates in how the loss scaling factor interacts with the $\mathbf{L_2}$ norm calculation for $\text{DP-SGD}$ clipping.

\paragraph{Mean-Aggregated Loss and Skew Amplification.}
The standard loss, $\ell_{i,t}^{\mathrm{mean}} = \frac{1}{d}\sum_{j=1}^d \ell_{i,t}^{(j)}$, produces a gradient vector $\mathbf{g}_{i,t}^{\mathrm{mean}}$ where every unscaled gradient component $\hat{\mathbf{g}}_k$ is uniformly scaled down by the factor $1/d$: $\mathbf{g}_k^{\mathrm{mean}} = \frac{1}{d} \cdot \hat{\mathbf{g}}_k$.
The per-sample clipping mechanism uses the $\mathbf{L_2}$ norm: $\|\mathbf{g}^{\mathrm{mean}}\|_2 = \sqrt{\sum_k (\mathbf{g}_k^{\mathrm{mean}})^2}$. This quadratic aggregation over scaled components leads to:
\begin{enumerate}
    \item \textbf{Suppression:} The $1/d$ factor pushes most gradient components below $1.0$. When squared, their contribution to $\|\mathbf{g}^{\mathrm{mean}}\|_2$ becomes negligible.
    \item \textbf{Outlier Dominance:} Given the \textbf{unequal feature variance} prevalent in mixed-type tabular data, only components corresponding to extreme errors ($\mathbf{g}_k^{\mathrm{mean}} \gg 1$) retain significant magnitude after squaring. This process quadratically enhances the relative influence of these few largest feature gradients.
\end{enumerate}
This dynamics results in a highly skewed distribution of $\|\mathbf{g}^{\mathrm{mean}}\|_2$ with a small mean but a heavy tail (Fig.\ref{fig:grad_norms}).

\paragraph{FA Loss and Relative Variability Reduction.}
The FA Loss, $\ell_{i,t}^{\mathrm{FA}} = \sum_{j=1}^d \ell_{i,t}^{(j)}$, uses unscaled gradients $\mathbf{g}^{\mathrm{FA}} = \hat{\mathbf{g}}$.
While the expected absolute magnitude $\mathbf{E}[\|\mathbf{g}^{\mathrm{FA}}\|_2]$ is larger, the distribution of the norm benefits from a profound concentration effect. The summation of independent (or weakly correlated) errors causes the resulting distribution of $\|\mathbf{g}^{\mathrm{FA}}\|_2$ to become more symmetric and better-behaved, aligning with the Central Limit Theorem.

This results in a significant reduction in the relative variability of the norm:
\[
\mathrm{Rel. Var.}(\|\mathbf{g}\|_2) = \frac{\mathbf{Var}(\|\mathbf{g}\|_2)}{\mathbf{E}[\|\mathbf{g}\|_2]^2}.
\]
This mechanism efficiently reduces the clipping-induced bias and stabilizes $\text{DP-SGD}$ updates, leading to the observed utility gains without altering the formal $(\varepsilon, \delta)$ guarantee.

\subsection{Adaptive Timestep (AT) Sampling}
\label{subsec:AT_sampling}

\paragraph{Objective and estimator.}
Let $u(t)=1/T$ be uniform over timesteps $\{1,\dots,T\}$ and define
\[
L(\theta)\;=\;\mathbb{E}_{t\sim u}\,\mathbb{E}_i\bigl[\ell_{i,t}(\theta)\bigr],\qquad
g_t(\theta)\;=\;\mathbb{E}_i\bigl[\nabla_\theta \ell_{i,t}(\theta)\bigr].
\]
To estimate $\nabla L(\theta)=\mathbb{E}_{t\sim u}[g_t(\theta)]$ by sampling $t\sim q(t)$, the unbiased importance-sampling (IS) estimator is
\[
G\;=\;\frac{u(t)}{q(t)}\cdot\frac{1}{B}\sum_{i\in B}\nabla_\theta \ell_{i,t}(\theta).
\]

\paragraph{Variance-minimizing $q$.}
Ignoring DP for the moment and treating $g_t$ as mean-zero centered fluctuations around its expectation, a standard IS calculation yields the second moment
\[
\mathbb{E}\|G\|_2^2\;=\;\sum_{t=1}^T \frac{u(t)^2}{q(t)}\, \mathbb{E}\Bigl\| \tfrac{1}{B}\textstyle\sum_{i\in B}\nabla \ell_{i,t}\Bigr\|_2^2
\;\propto\; \sum_t \frac{u(t)^2}{q(t)}\, v(t),
\]
where $v(t):=\mathbb{E}\|\nabla \ell_{i,t}\|_2^2/B$ collects per-timestep gradient energy (constants in $B$ omitted for clarity). Minimizing $\sum_t \frac{u(t)^2}{q(t)}v(t)$ subject to $\sum_t q(t)=1$ gives by Cauchy–Schwarz (or convex duality)
\[
q^\star(t)\;\propto\; u(t)\,\sqrt{v(t)}\;\propto\; u(t)\,\sqrt{\mathbb{E}\|\nabla \ell_{i,t}\|_2^2}\,.
\tag{A1}
\label{eq:optq-nodp}
\]

\paragraph{DP-aware surrogate.}
Under DP-SGD we use clipped per-example gradients $g^{\mathrm{clip}}_{i,t}=g_{i,t}\cdot \min(1,\,C/\|g_{i,t}\|_2)$ before adding noise. A conservative DP-aware surrogate for per-timestep signal is
\[
s(t)\;:=\;\mathbb{E}\bigl[\|g^{\mathrm{clip}}_{i,t}\|_2\bigr]
\;=\;\mathbb{E}\!\left[\min\!\bigl(\|g_{i,t}\|_2,\,C\bigr)\right]\in[0,C].
\]
Replacing $\sqrt{\mathbb{E}\|\nabla \ell_{i,t}\|_2^2}$ in \eqref{eq:optq-nodp} by $s(t)$ yields the DP-aware target
\[
q^\star_{\mathrm{DP}}(t)\;\propto\; u(t)\, s(t)\;\propto\; s(t),
\tag{A2}
\label{eq:optq-dp}
\]
which prioritizes timesteps with large \emph{post-clipping} gradient magnitude and de-emphasizes those dominated by DP noise.

\paragraph{Parametric proxy and annealing.}
Directly estimating $s(t)$ online is expensive and noisy. Empirically in diffusion training with DP, the function $s(t)$ is (i) small near the pure-noise tail (very large $t$), (ii) grows as $t$ decreases toward informative denoising regimes, and (iii) its mode shifts during learning (early epochs favoring larger $t$, later epochs smaller $t$). We approximate $q^\star_{\mathrm{DP}}(t)$ by the power-law family
\[
P_k(t)\;\propto\; t^{\alpha_k}, \qquad
\alpha_k=\alpha_{\text{start}} + \tfrac{k}{K}\bigl(\alpha_{\text{end}}-\alpha_{\text{start}}\bigr),
\]
with $\alpha_{\text{start}}>0$ and $\alpha_{\text{end}}<0$. This one-parameter schedule smoothly transfers probability mass from later (large-$t$) to earlier (small-$t$) denoising stages as training progresses. 

\begin{prop}[Power-law proxy tracks $q^\star_{\mathrm{DP}}$]
Assume that for epoch $k$ the DP-effective signal $s_k(t)$ is monotone in $t$ (decreasing after an initial stabilization period). Then there exists $\alpha_k\in\mathbb{R}$ such that $t^{\alpha_k}$ fits $s_k(t)$ in the least-squares sense on $\{\log t\}$, i.e., $\alpha_k=\arg\min_\alpha \sum_t \bigl(\log s_k(t) - \alpha \log t - c\bigr)^2$ for some $c$. Consequently, $P_k(t)\propto t^{\alpha_k}$ approximates $q^\star_{\mathrm{DP}}(t)\propto s_k(t)$ up to a normalization constant.
\end{prop}


\paragraph{With or without IS correction.}
One may include IS correction factor $u(t)/P_k(t)$ to keep an unbiased estimator for $\nabla L(\theta)$. In practice, diffusion training commonly \emph{omits} this correction and directly optimizes the weighted objective $L_{P_k}(\theta)=\mathbb{E}_{t\sim P_k}\mathbb{E}_i[\ell_{i,t}]$, which emphasizes DP-informative timesteps and empirically improves convergence. Our results use the latter for stability.

\paragraph{Privacy unchanged.}
AT modifies only the sampling distribution over $t$; it does not change per-example clipping, noise scale, sampling rate $q$ (in the DP sense), or the accountant. Therefore the $(\varepsilon,\delta)$ computed by PRV (or RDP) is unaffected.

\paragraph{Takeaway.}
AT is a DP-aware variance-reduction mechanism: it reallocates training effort toward timesteps with larger post-clipping gradient magnitude, using a smooth power-law schedule that tracks the evolving signal across training while preserving the formal privacy guarantee.

\section{Differential Privacy Enforcement}
\noindent\textit{DP-SGD + PRV Accounting.}  
Gradient clipping and Gaussian noise injection follow the standard DP-SGD framework. We select clipping norm \(C\) and noise scale \(\sigma\) to meet a target \((\varepsilon, \delta)\). The PRV accountant composes per-step privacy loss for final guarantee. Privacy accounting uses the Privacy Random Variable (PRV)~\cite{gopi2021numerical} method to compose over steps.  

\begin{prop}
Let \(\theta\) be the parameters learned by DP-FinDiff after \(K\) steps with batch size \(B\), clip norm \(C\), and noise scale \(\sigma\). Then \(\theta\) satisfies \((\varepsilon,\delta)\)-differential privacy, where \(\varepsilon\) is obtained via the PRV accountant over those parameters. The FA loss and AT sampler do not alter this guarantee, as they only modify the loss structure and sampling distribution, not the privacy mechanism.
\end{prop}  

\textbf{Discussion of Privacy–Utility Trade-off.}
The embedding-based categorical encoding directly mitigates encoding overhead and aids scalability, because the model handles high-cardinality features in a lower-dimensional latent space. Meanwhile, FA and AT are designed to reduce the effective noise impact on gradient updates, thus addressing utility loss by preserving more signal under the same \((\varepsilon, \delta)\). Together, these design choices guide DP-FinDiff toward improved privacy–utility balances in sensitive tabular domains.

\section{Hyperparameter Sensitivity Analysis.}
\begin{figure}[h!]
    \centering
    \includegraphics[width=0.8\linewidth]{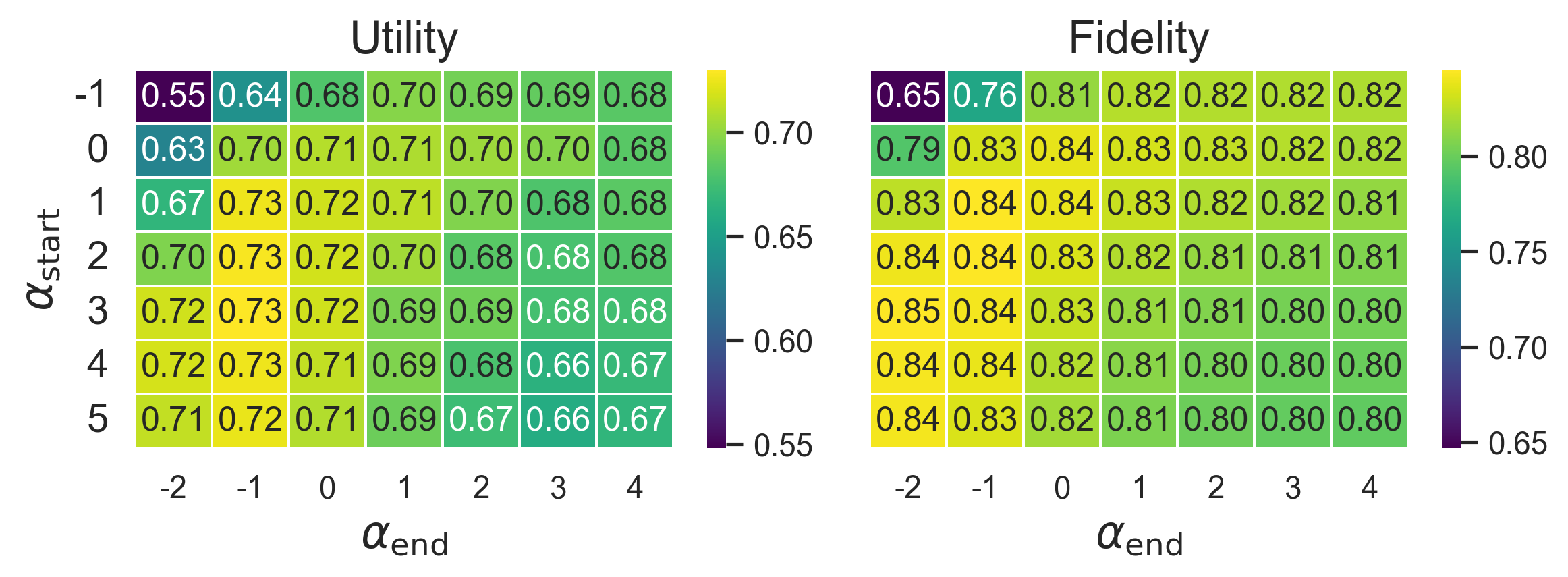}
    \caption{Evaluation of the \textit{adaptive timestep} sampling hyperparameters. The results are averaged across all datasets and all DP levels ($\varepsilon \in$ \{0.2, 1, 10\}). Largest performance improvements are observed at $\alpha_{\text{start}}$ $\leq$ 3 and $\alpha_{\text{end}}$ = -1.}
  \label{fig:AT_grid}
\end{figure}

We evaluate the effect of the adaptive timestep sampler’s hyperparameters $\alpha_\text{start}$ and $\alpha_\text{end}$ on model performance, averaged over all datasets and privacy levels $\varepsilon \in$ \{0.2, 1, 10\}, as shown on~\autoref{fig:AT_grid}. Utility and fidelity are both sensitive to these parameters, with the best results observed when $\alpha_\text{start} \leq$ 3 and $\alpha_\text{end}$ = -1. This setting reflects an effective sampling schedule that initially emphasizes later timesteps, where gradients are smaller and more stable under DP and gradually shifts toward earlier steps for fine-grained synthesis. Extreme values (e.g., large $\alpha_\text{start}$ or positive $\alpha_\text{end}$) degrade performance, confirming the importance of carefully controlling the temporal focus during training.

\section{Qualitative Evaluation.}
\begin{figure}[t!]
  \centering

  \begin{subfigure}[b]{0.19\textwidth}
    \centering
    \includegraphics[width=\linewidth, ]{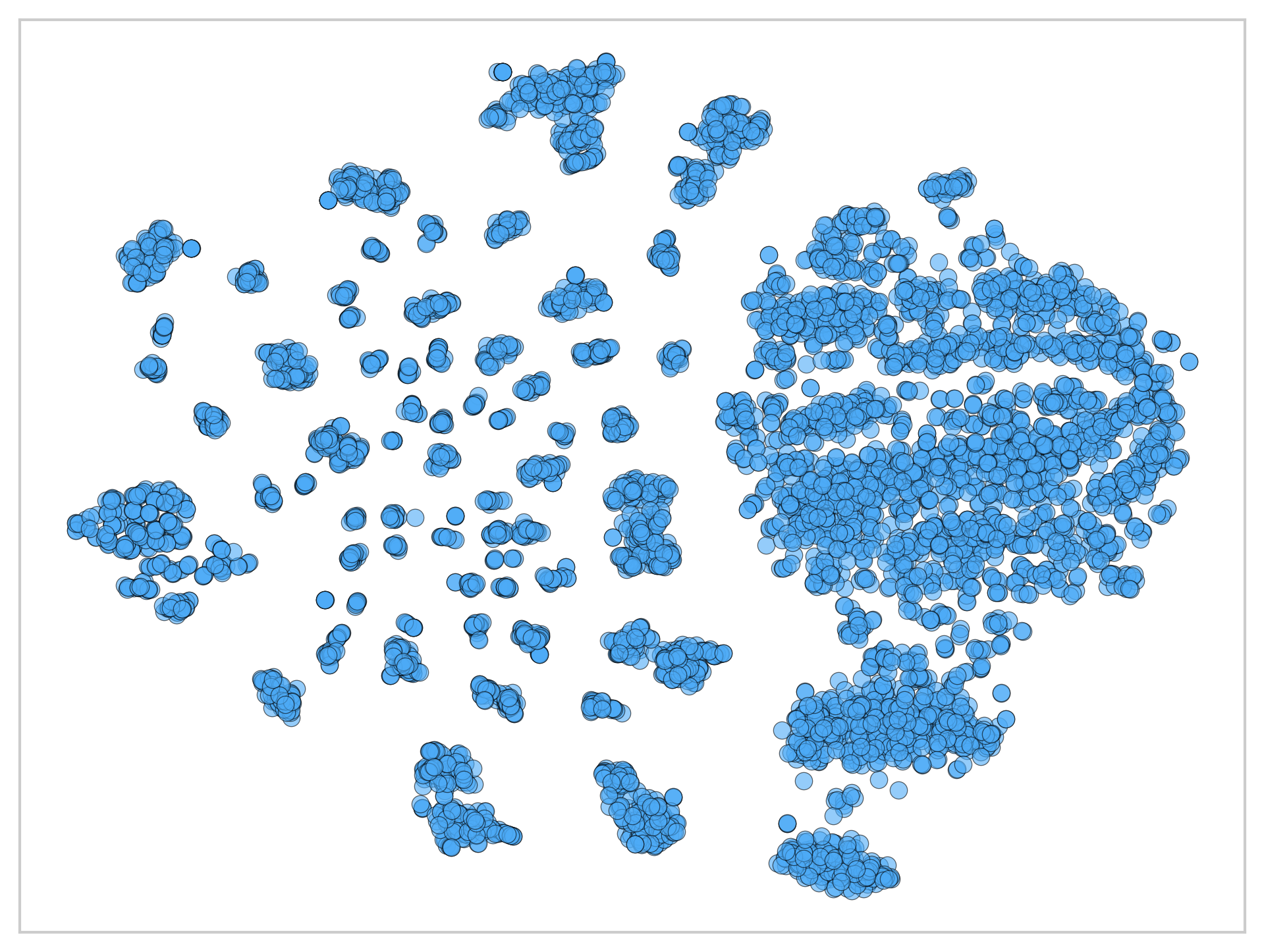}
    \caption{Original}
    \label{fig:tsne_real}
  \end{subfigure}
  \hfill
  \begin{subfigure}[b]{0.19\textwidth}
    \centering
    \includegraphics[width=\linewidth, ]{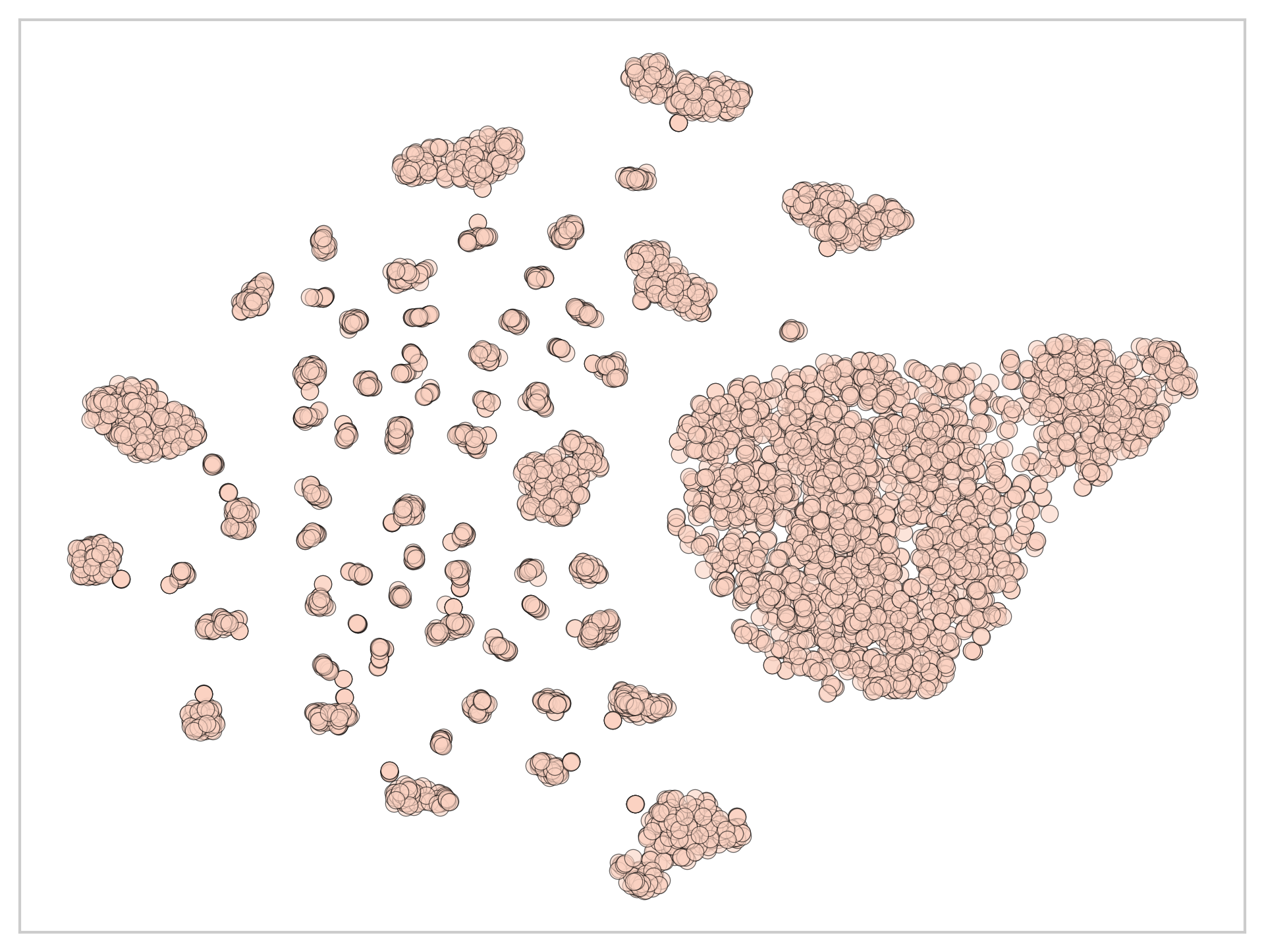}
    \caption{Synthetic ($\varepsilon$=$\infty$)}
    \label{fig:tsne_noDP}
  \end{subfigure}
  \hfill
  \begin{subfigure}[b]{0.19\linewidth}
    \centering
    \includegraphics[width=\linewidth, ]{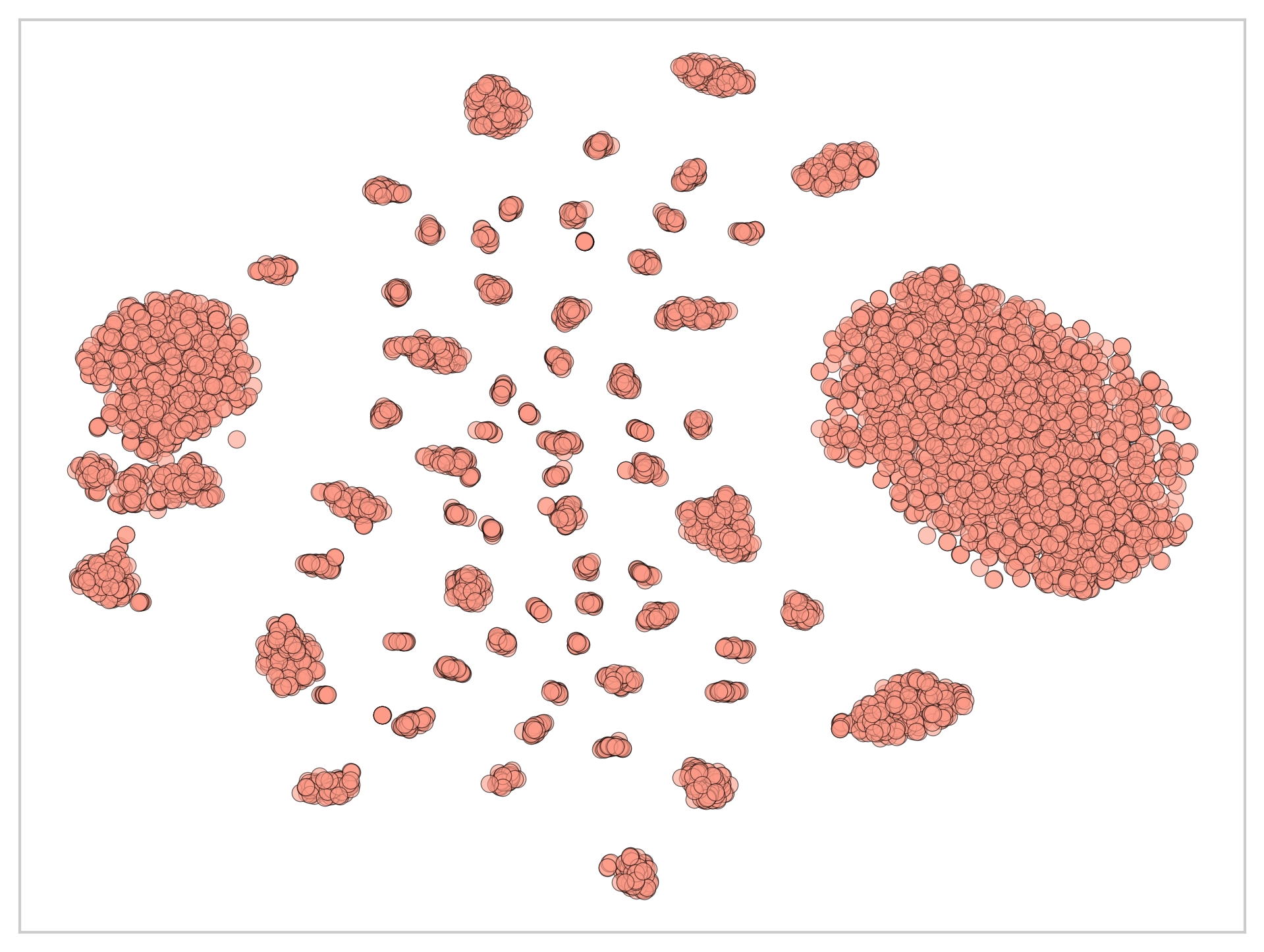}
    \caption{Synthetic ($\varepsilon$=10)}
    \label{fig:tsne_DP10}
  \end{subfigure}
    \hfill
  \begin{subfigure}[b]{0.19\linewidth}
    \centering
    \includegraphics[width=\linewidth, ]{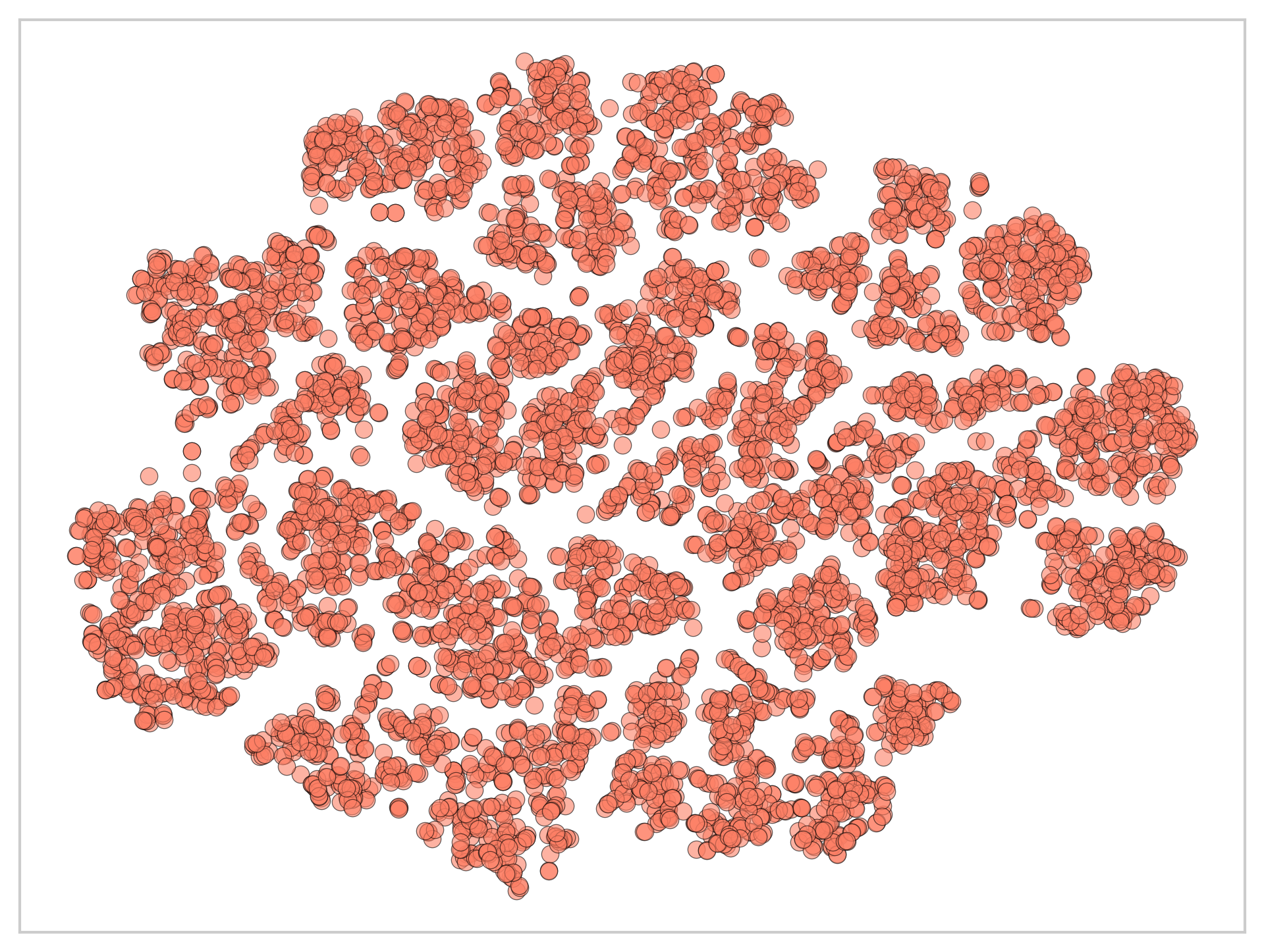}
    \caption{Synthetic ($\varepsilon$=1)}
    \label{fig:tsne_DP1}
  \end{subfigure}
    \hfill
  \begin{subfigure}[b]{0.19\linewidth}
    \centering
    \includegraphics[width=\linewidth, ]{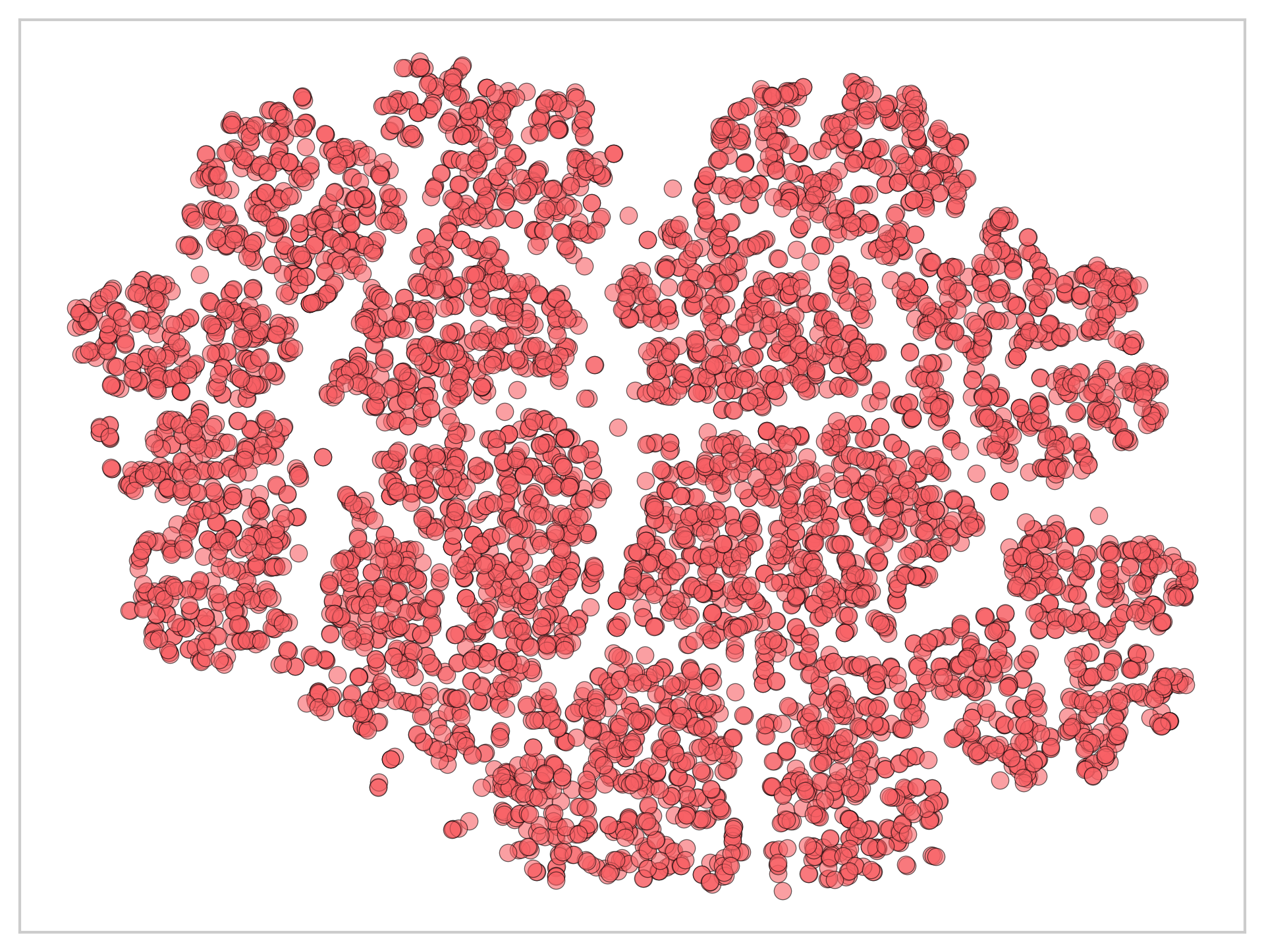}
    \caption{Synthetic ($\varepsilon$=0.2)}
    \label{fig:tsne_DP02}
  \end{subfigure}

  \begin{subfigure}[b]{0.19\textwidth}
    \centering
    \includegraphics[width=\linewidth, ]{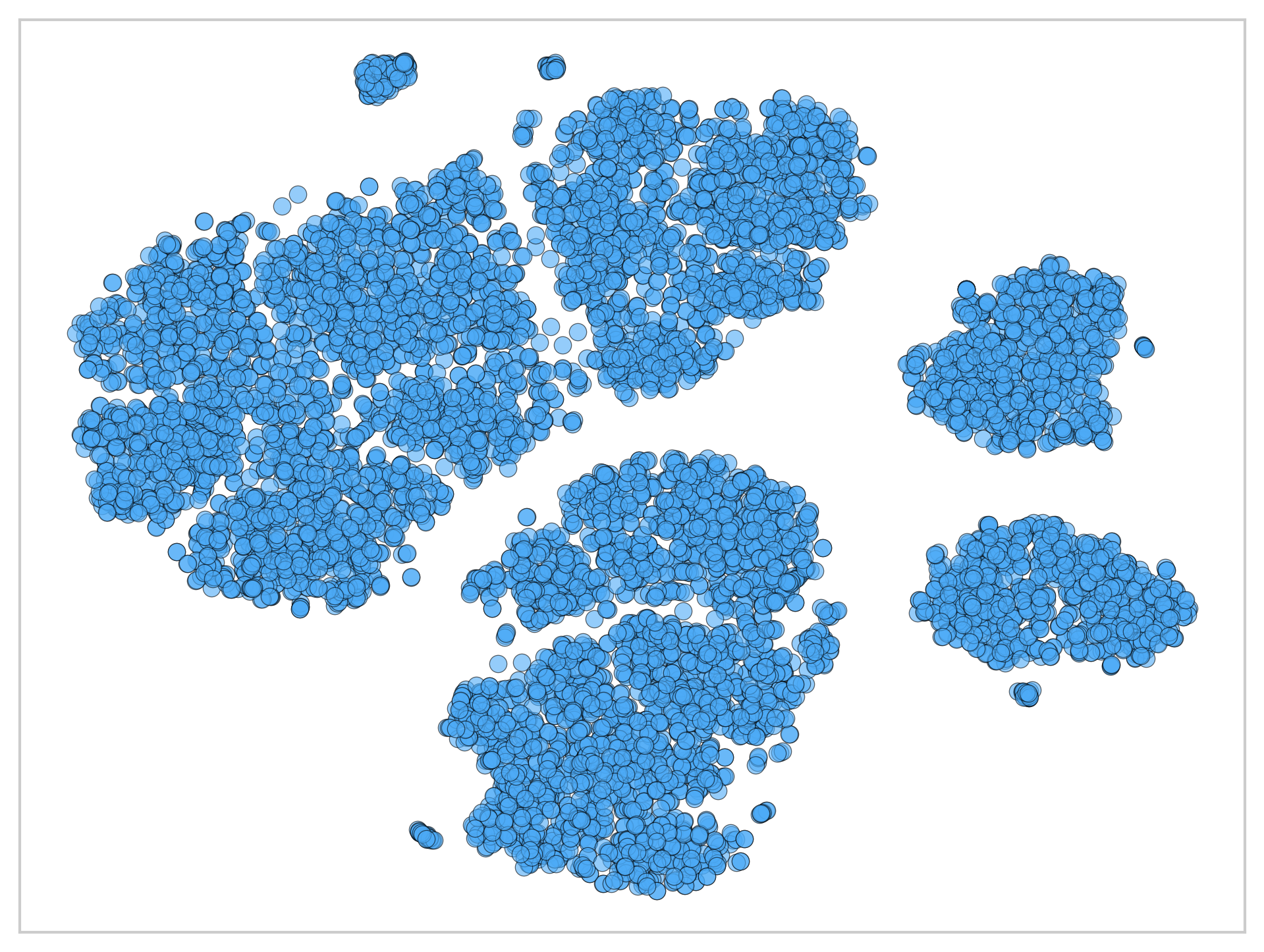}
    \caption{Original}
    \label{fig:tsne_real}
  \end{subfigure}
  \hfill
  \begin{subfigure}[b]{0.19\textwidth}
    \centering
    \includegraphics[width=\linewidth, ]{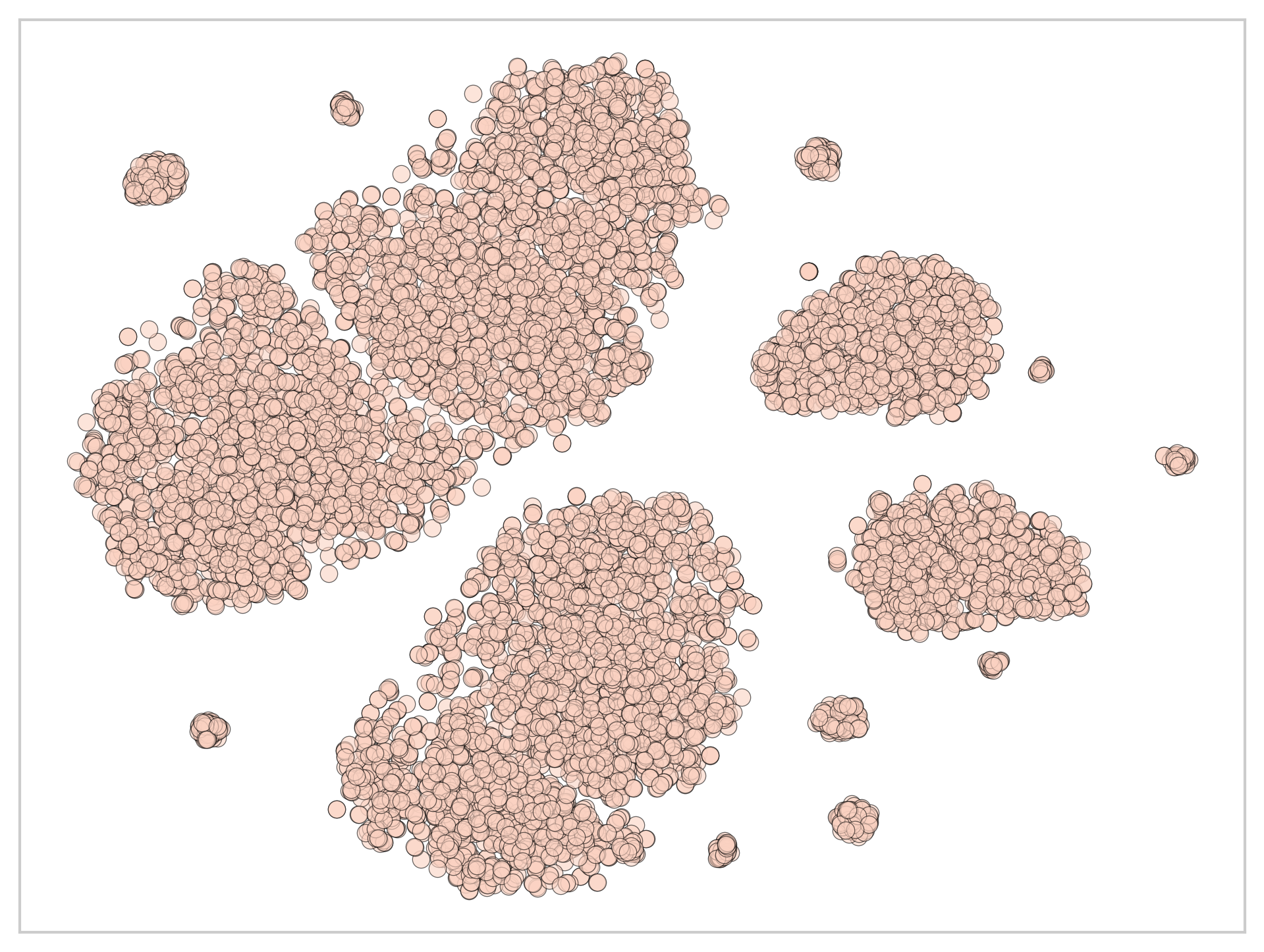}
    \caption{Synthetic ($\varepsilon$=$\infty$)}
    \label{fig:tsne_noDP}
  \end{subfigure}
  \hfill  
  \begin{subfigure}[b]{0.19\linewidth}
    \centering
    \includegraphics[width=\linewidth, ]{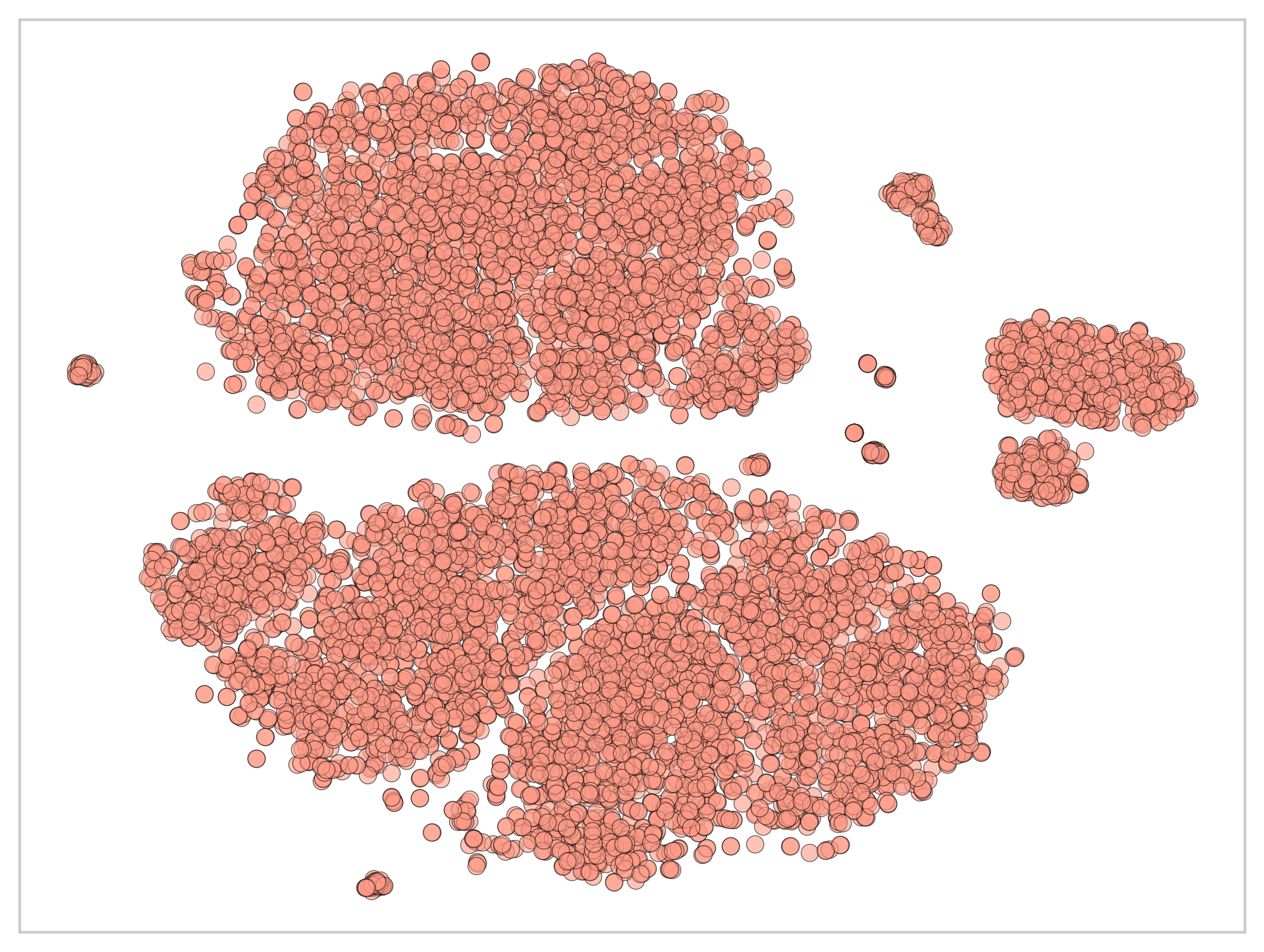}
    \caption{Synthetic ($\varepsilon$=10)}
    \label{fig:tsne_DP10}
  \end{subfigure}
    \hfill
  \begin{subfigure}[b]{0.19\linewidth}
    \centering
    \includegraphics[width=\linewidth, ]{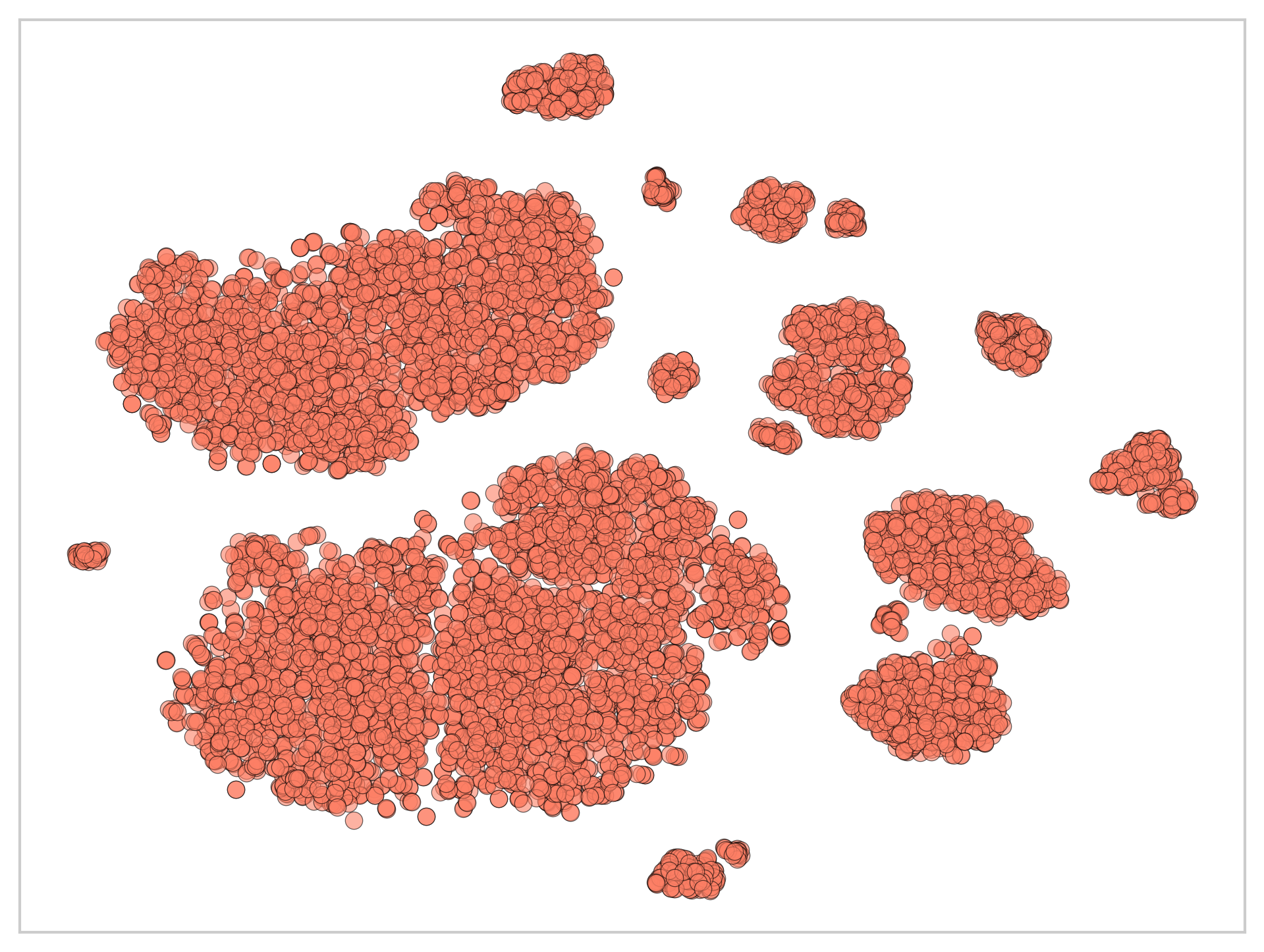}
    \caption{Synthetic ($\varepsilon$=1)}
    \label{fig:tsne_DP1}
  \end{subfigure}
    \hfill
  \begin{subfigure}[b]{0.19\linewidth}
    \centering
    \includegraphics[width=\linewidth, ]{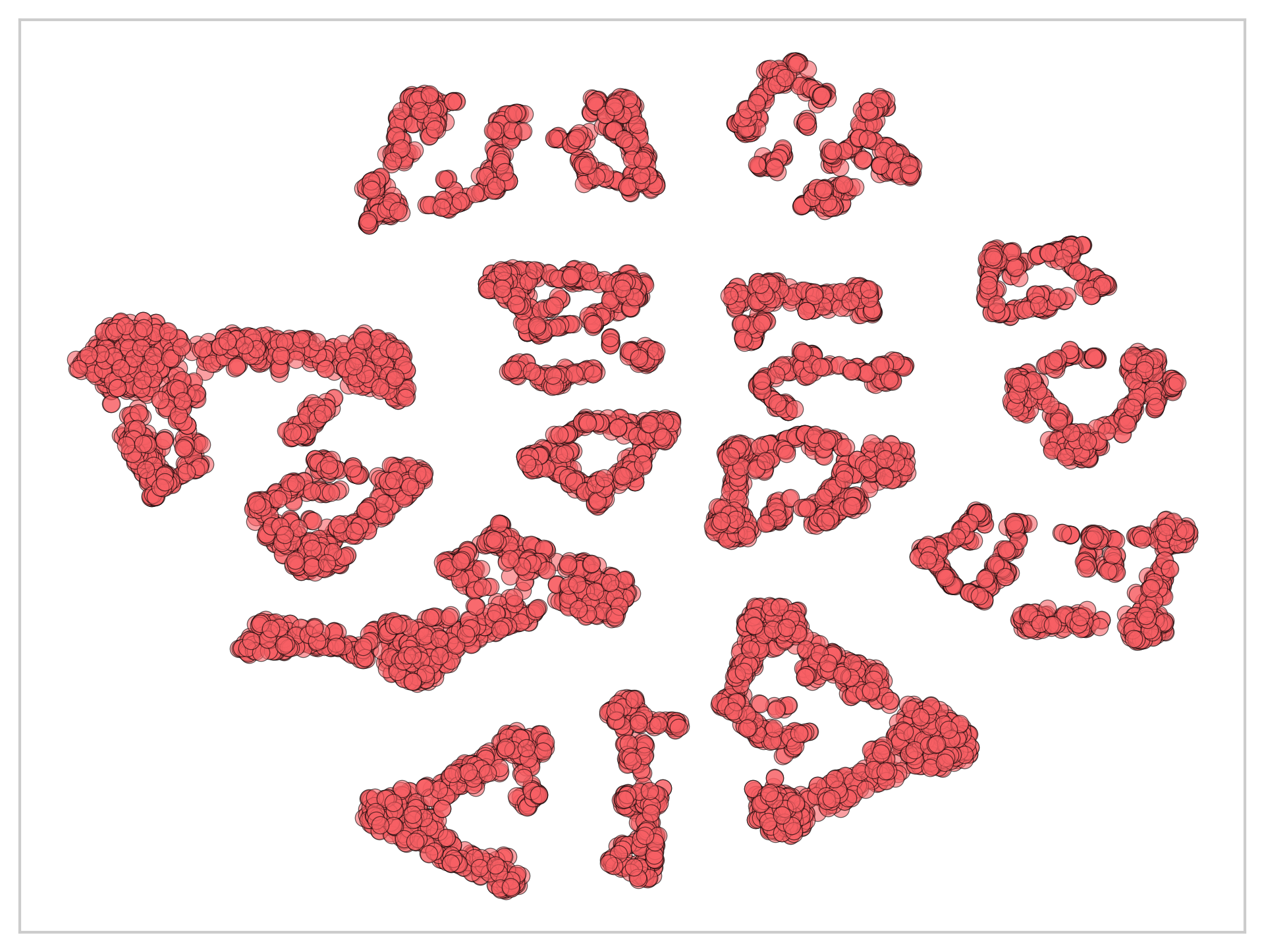}
    \caption{Synthetic ($\varepsilon$=0.2)}
    \label{fig:tsne_DP02}
  \end{subfigure}

  \caption{t-SNE visualization of Credit (top row) and Marketing (bottom row) datasets from \textit{DP-FinDiff} under various DP levels. Shown are: original data (a, f), synthetic data without DP (b, g), and with DP $\varepsilon \in$ \{10, 1, 0.2\} (c-e, h-j) . As privacy level increases (reflected by a decrease in $\varepsilon$), the structural integrity of the synthetic data diminishes, resulting from the increased DP noise.}
  \label{fig:TSNE}
\end{figure}

We use t-SNE to visualize the structural fidelity of synthetic data generated by \textit{DP-FinDiff} under varying privacy budgets $\varepsilon \in$ \{$\infty$, 10, 1, 0.2\} (\autoref{fig:TSNE}). These plots compare the original and synthetic distributions to assess the impact of DP noise on data structure. \textit{DP-FinDiff} preserves the global cluster structure across all datasets. At $\varepsilon$=$\infty$, synthetic samples closely match the original distribution. As $\varepsilon$ decreases, mild distortions emerge, particularly at $\varepsilon$=0.2, where cluster fragmentation becomes more apparent. These patterns mirror the privacy-utility tradeoffs seen in quantitative metrics.
Overall, \textit{DP-FinDiff} maintains structural patterns, supporting its applicability to financial tasks like fraud detection that depend on preserving cluster-level semantics.

\end{document}